\documentclass[10pt,twocolumn,letterpaper]{article}

\usepackage[pagenumbers]{cvpr} 

\usepackage{graphicx}
\usepackage{amsmath}
\usepackage{amssymb}
\usepackage{booktabs}

\usepackage{subcaption}

\usepackage{sg-macros}
\usepackage{tabularx} 
\newcolumntype{C}{>{\centering\arraybackslash}X}
\usepackage{booktabs}
\usepackage[pagebackref,breaklinks,colorlinks]{hyperref}

\usepackage[capitalize]{cleveref}
\crefname{section}{Sec.}{Secs.}
\Crefname{section}{Section}{Sections}
\Crefname{table}{Table}{Tables}
\crefname{table}{Tab.}{Tabs.}

\def\cvprPaperID{2752} 
\def\confName{CVPR}
\def\confYear{2022}

\newsavebox{\measurebox}

\definecolor{codegreen}{rgb}{0,0.6,0}
\definecolor{codegray}{rgb}{0.5,0.5,0.5}
\definecolor{codepurple}{rgb}{0.58,0,0.82}
\definecolor{backcolour}{rgb}{1.0,1.0,1.0}

\lstdefinestyle{mystyle}{
    backgroundcolor=\color{backcolour},   
    commentstyle=\color{codegreen},
    frame=none,
    language=Python,
    keywordstyle=\color{magenta},
    numberstyle=\scriptsize\color{codegray},
    stringstyle=\color{codepurple},
    basicstyle=\ttfamily\footnotesize,
    breakatwhitespace=false,         
    breaklines=true,                 
    captionpos=b,                    
    keepspaces=true,                 
    numbers=none,                    
    numbersep=5pt,                  
    showspaces=false,                
    showstringspaces=false,
    showtabs=false,                  
    tabsize=2
}

\begin{document}

\title{Efficient Two-Stage Detection of Human--Object Interactions\\with a Novel Unary--Pairwise Transformer}

\author{Frederic Z. Zhang$^{1, 3}$ \quad Dylan Campbell$^{2, 3}$ \quad Stephen Gould$^{1, 3}$ \\
$^1$The Australian National University \quad $^2$University of Oxford \\ $^3$Australian Centre for Robotic Vision\\
{\tt\small \href{https://fredzzhang.com/unary-pairwise-transformers}{https://fredzzhang.com/unary-pairwise-transformers}}
}

\newcommand{\fz}[1]{\comment{FZ}{#1}}
\newcommand{\sg}[1]{\comment{SG}{#1}}
\newcommand{\dc}[1]{\comment{DC}{#1}}

\maketitle

\begin{abstract}
Recent developments in transformer models for visual data have led to significant improvements in recognition and detection tasks. In particular, using learnable queries in place of region proposals has given rise to a new class of one-stage detection models, spearheaded by the Detection Transformer (DETR). Variations on this one-stage approach have since dominated human--object interaction (HOI) detection. However, the success of such one-stage HOI detectors can largely be attributed to the representation power of transformers. We discovered that when equipped with the same transformer, their two-stage counterparts can be more performant and memory-efficient, while taking a fraction of the time to train. In this work, we propose the Unary--Pairwise Transformer, a two-stage detector that exploits unary and pairwise representations for HOIs. We observe that the unary and pairwise parts of our transformer network specialise, with the former preferentially increasing the scores of positive examples and the latter decreasing the scores of negative examples. We evaluate our method on the HICO-DET and V-COCO datasets, and significantly outperform state-of-the-art approaches. At inference time, our model with ResNet50 approaches real-time performance on a single GPU.
\end{abstract}

\section{Introduction}

\begin{figure}[t]
    \begin{subfigure}[t]{\linewidth}
    \centering
    \includegraphics[width=0.6\linewidth]{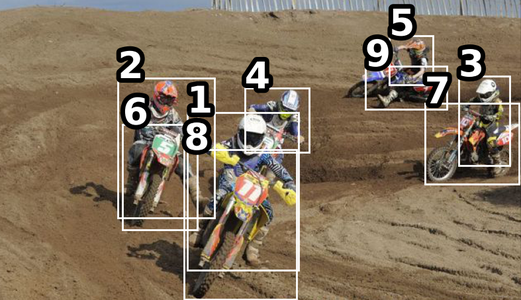}
    \caption{Image with human and object detections.}
    \label{fig:teaser-sample}
\end{subfigure}
\begin{subfigure}[t]{\linewidth}
    \centering
    \includegraphics[width=\linewidth]{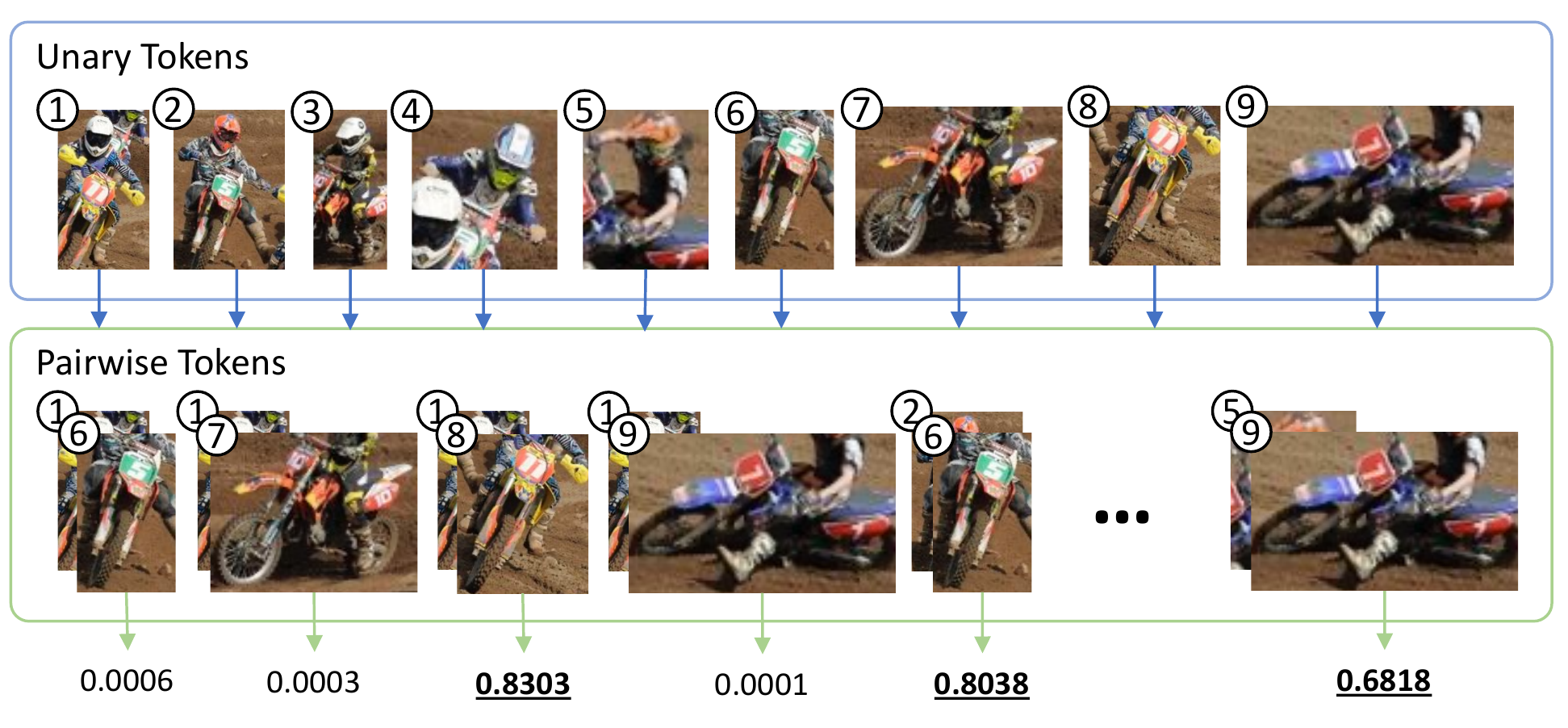}
    \caption{Unary and pairwise tokens with predicted scores (\emph{riding a motorcycle}).}
    \label{fig:teaser-tokens}
\end{subfigure}
\caption{Our Unary--Pairwise Transformer encodes human and object instances individually and in pairs, allowing it to reason about the data in complementary ways. In this example, our network correctly identifies the interactive pairs for the action \textit{riding a motorcycle}, while suppressing the visually-similar non-interactive pairs and those with different associated actions.
}
\label{fig:teaser}
\vspace{-10pt}
\end{figure}

Human--object interaction (HOI) detectors localise interactive human--object pairs in an image and classify the actions. They can be categorised as one- or two-stage, mirroring the grouping of object detectors.
Exemplified by Faster R-CNN~\cite{fasterrcnn}, two-stage object detectors typically include a region proposal network, which explicitly encodes potential regions of interest in the form of bounding boxes. These bounding boxes can then be classified and further refined via regression in a downstream network. In contrast, one-stage detectors, such as RetinaNet~\cite{retinanet}, retain the abstract feature representations of objects throughout the network, and decode them into bounding boxes and classification scores at the end of the pipeline.

\begin{figure}[t]
    \centering
    \includegraphics[width=\linewidth]{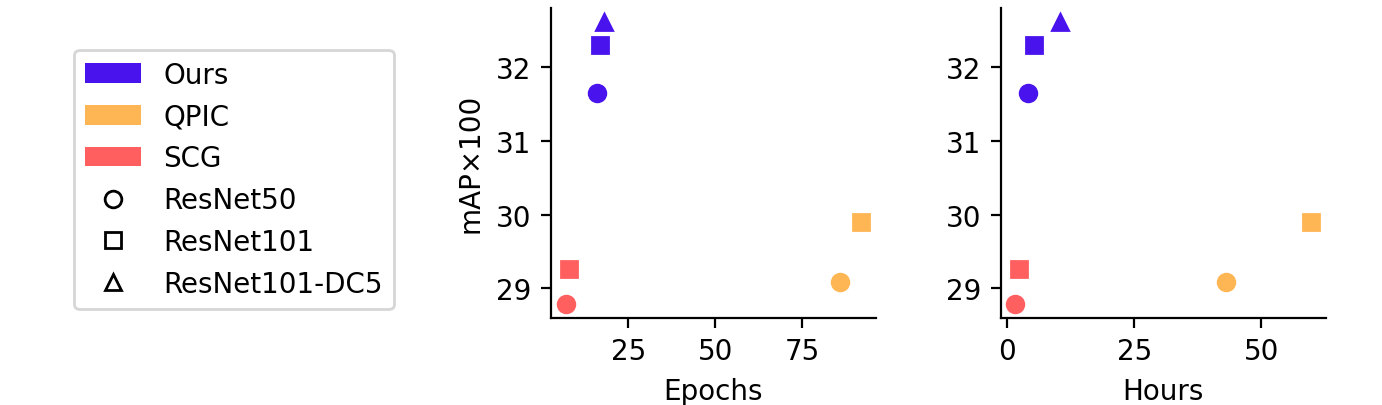}
    \caption{Mean average precision as a function of the number of epochs (left) and training time (right) to convergence. The backbone networks for all methods have been initialised with the same weights and trained on 8 GeForce GTX TITAN X GPUs.}
    \label{fig:convg-time}
\end{figure}

In addition to the same categorisation convention, HOI detectors need to localise two bounding boxes per instance instead of one. Early works~\cite{hicodet,gpnn,no-frills,tin} employ a pre-trained object detector to obtain a set of human and object boxes, which are paired up exhaustively and processed by a downstream network for interaction classification. This methodology coincides with that of two-stage detectors and quickly became the mainstream approach due to the accessibility of high-quality pre-trained object detectors.
The first instance of one-stage HOI detectors was introduced by Liao \etal.~\cite{ppdm}. They characterised human--object pairs as interaction points, represented as the midpoint of the human and object box centres. Recently, due to the great success in using learnable queries in transformer decoders for localisation~\cite{detr}, the development of one-stage HOI detectors has been greatly advanced. However, HOI detectors that adapt the DETR model rely heavily on the transformer, which is notoriously difficult to train~\cite{train-xfmer}, to produce discriminative features. In particular, when initialised with DETR's pre-trained weights, the decoder attends to regions of high objectness by default. The heavy-weight decoder stack then has to be adapted to attend to regions of high interactiveness. Consequently, training such one-stage detectors often consumes large amounts of memory and time as shown in \cref{fig:convg-time}.
In contrast, two-stage HOI detectors do not repurpose the backbone network, but maintain it as an object detector. Since the first half of the pipeline already functions as intended at the beginning of training, the second half can be trained quickly for the specific task of HOI detection. Furthermore, since the object detector can be decoupled from the downstream interaction head during training, its weights can be frozen, and a lighter-weight network can be used for interaction detection, saving a substantial amount of memory and computational resources.

\begin{table}[t]\small
    \vspace{-4pt}
    \caption{The performance discrepancy between existing state-of-the-art one-stage and two-stage HOI detectors is largely attributable to the choice of backbone network. We report the mean average precision ($\times 100$) on the HICO-DET~\cite{hicodet} test set.}
    \label{tab:one-vs-two}
    \setlength{\tabcolsep}{6pt} 
    \vspace{-4pt}
    \begin{tabularx}{\linewidth}{l l l C}
        \toprule
        \textbf{Method} & \textbf{Type} & \textbf{Detector Backbone} & \textbf{mAP} \\
        \midrule
        SCG~\cite{scg} & two-stage & Faster R-CNN R-50-FPN & 24.88 \\
        SCG~\cite{scg} & two-stage & DETR R-50 & 28.79 \\
        SCG~\cite{scg} & two-stage & DETR R-101 & \textbf{29.26} \\
        \midrule
        QPIC~\cite{qpic} & one-stage & DETR R-50 & 29.07 \\
        QPIC~\cite{qpic} & one-stage & DETR R-101 & \textbf{29.90} \\
    \midrule
    Ours & two-stage & DETR R-50 & 31.66 \\
    Ours & two-stage & DETR R-101 & \textbf{32.31} \\
        \bottomrule
    \end{tabularx}
    \vspace{-6pt}
\end{table}

Despite these advantages, the performance of two-stage detectors has lagged behind their one-stage counterparts. However, most of these two-stage models used Faster R-CNN~\cite{fasterrcnn} rather than more recent object detectors. We found that simply replacing Faster R-CNN with the DETR model in an existing two-stage detector (SCG)~\cite{scg} resulted in a significant improvement, putting it on par with a state-of-the-art one-stage detector (QPIC), as shown in \cref{tab:one-vs-two}. We attribute this performance gain to the representation power of transformers and bipartite matching loss~\cite{detr}. The latter is particularly important because it resolves the misalignment between the training procedure and evaluation protocol. The evaluation protocol dictates that, amongst all detections associated with the same ground truth, the highest scoring one is the true positive while the others are false positives. Without bipartite matching, all such detections will be labelled as positives. The detector then has to resort to heuristics such as non-maximum suppression to mitigate the issue, resulting in procedural misalignment.

We propose a two-stage model that refines the output features from DETR with additional transformer layers for HOI classification. As shown in \cref{fig:teaser}, we encode the instance information in two ways: a unary representation where individual human and object instances are encoded separately, and a pairwise representation where human--object pairs are encoded jointly. These representations provide orthogonal information, and we observe different behaviours in their associated layers. The unary encoder layer preferentially increases the predicted interaction scores for positive examples, while the pairwise encoder layer suppresses the negative examples. As a result, this complementary behaviour widens the gap between scores of positive and negative examples, particularly benefiting ranking metrics such as mean average precision (mAP).

Our primary contribution is a novel and efficient two-stage HOI detector with unary and pairwise encodings. Our secondary contribution is demonstrating how pairwise box positional encodings---critical for HOI detection---can be incorporated into a transformer architecture, enabling it to jointly reason about unary appearance and pairwise spatial information. We further provide a detailed analysis on the behaviour of the two encoder layers, showing that they have complementary properties. Our proposed model not only outperforms state-of-the-art methods, but also consumes much less time and memory to train. The latter allows us to employ more memory-intensive backbone networks, further improving the performance.

\begin{figure*}[t]
\centering
\includegraphics[width=\linewidth]{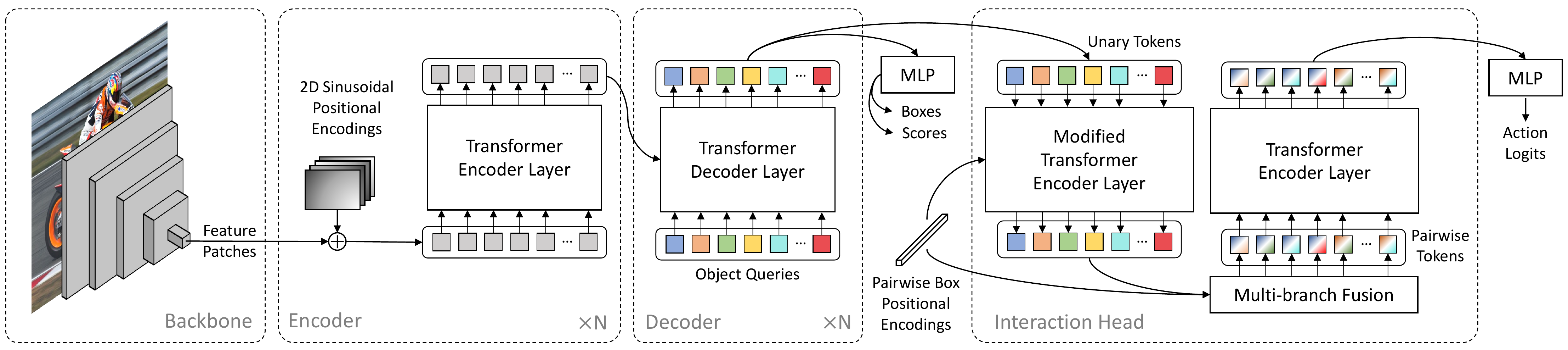}
\caption{Flowchart for our unary--pairwise transformer. An input image is processed by a backbone CNN to produce image features, which are partitioned into patches of equal size and augmented with sinusoidal positional encodings. These tokens are fed into the DETR~\cite{detr} transformer encoder--decoder stack, generating new features for a fixed number of learnable object queries. These are decoded by an MLP as object classification scores and bounding boxes, and are also passed to the interaction head as unary tokens. The interaction head also receives pairwise positional encodings computed from the predicted bounding box coordinates. A modified transformer encoder layer then refines the unary tokens using the pairwise positional encodings. The output tokens are paired up and fused with the same positional encodings to produce pairwise tokens, which are processed by a standard transformer encoder layer before an MLP decodes the final features as action classification scores.
}
\label{fig:diagram}
\end{figure*}

\section{Related work}

Transformer networks~\cite{xfmer}, initially developed for machine translation, have recently become ubiquitous in computer vision due to their representation power, flexibility, and global receptive field via the attention mechanism. The image transformer ViT~\cite{vit} represented an image as a set of spatial patches, each of which was encoded as a token through simple linear transformations. This approach for tokenising images rapidly gained traction and inspired many subsequent works~\cite{swint}. Another key innovation of transformers is the use of learnable queries in the decoder, which are initialised randomly and updated through alternating self-attention and cross-attention with encoder tokens. Carion \etal~\cite{detr} use these as object queries in place of conventional region proposals for their object detector. Together with a bipartite matching loss, this design gave rise to a new class of one-stage detection models that formulate the detection task as a set prediction problem. It has since inspired numerous works in HOI detection~\cite{qpic, hoitrans, hotr, asnet}.

To adapt the DETR model to HOI detection, Tamura \etal~\cite{qpic} and Zou \etal~\cite{hoitrans} add additional heads to the transformer in order to localise both the human and object, as well as predict the action. As for bipartite matching, additional cost terms are added for action prediction. On the other hand, Kim \etal~\cite{hotr} and Chen \etal~\cite{asnet} propose an interaction decoder to be used alongside the DETR instance decoder. It is specifically responsible for predicting the action while also matching the interactive human--object pairs. These aforementioned one-stage detectors have achieved tremendous success in pushing the state-of-the-art performance. However, they all require significant resources to train the models. In contrast, this work focuses on exploiting novel ideas to produce equally discriminative features while preserving the memory efficiency and low training time of two-stage detectors.

Two-stage HOI detectors have also undergone significant development recently. Li \etal~\cite{idn} studied the integration and decomposition of HOIs in an analogy to the superposition of waves in harmonic analysis. Hou \etal explored few-shot learning by fabricating object representations in feature space~\cite{fcl} and learning to transfer object affordance~\cite{atl}. Finally, Zhang \etal~\cite{scg} proposed to fuse features of different modalities within a graphical model to produce more discriminative features. We make use of this modality fusion in our transformer model and show that it leads to significant improvements.

\section{Unary--pairwise transformers}

To leverage the success of transformer-based detectors, we use DETR~\cite{detr} as our backbone object detector and focus on designing an effective and efficient interaction head for HOI detection, as shown in \cref{fig:diagram}. The interaction head consists of two types of transformer encoder layers, with the first layer modified to accommodate additional pairwise input. The first layer operates on unary tokens, \ie, individual human and object instances, while the second layer operates on pairwise tokens, \ie, human--object pairs. Based on our analysis and experimental observations in \cref{sec:macro} and \cref{sec:micro}, self-attention in the unary layer preferentially increases the interaction scores for positive HOI pairs, whereas self-attention in the pairwise layer decreases the scores for negative pairs. As such, we refer to these layers as \textit{cooperative} and \textit{competitive} layers respectively.

\subsection{Cooperative layer}
\label{sec:coop}

A standard transformer encoder layer takes as input a set of tokens and performs self-attention. Positional encodings are usually indispensable to compensate for the lack of order in the token set. Typically, sinusoidal functions of the position~\cite{xfmer} or learnable embeddings~\cite{detr} are used for this purpose. It is possible to extend sinusoidal encodings to bounding box coordinates, however, our unary tokens already contain positional information, since they were decoded into bounding boxes. Instead, we take this as an opportunity to inject pairwise spatial information into the transformer, something that has been shown to be helpful for the task of HOI detection~\cite{scg}. Specifically, we compute the unary and pairwise spatial features used by Zhang \etal~\cite{scg} from the bounding boxes, including the unary box centre, width and height, and pairwise intersection-over-union, relative area, and direction, and pass this through an MLP to obtain the pairwise positional encodings. We defer the full details to~\cref{app:pe}.
We also found that the usual additive approach did not perform as well for our positional encodings. So we slightly modified the attention operation in the transformer encoder layer to allow directly injecting the pairwise positional encodings into the computation of values and attention weights.

\begin{figure}[t]
    \centering
    \includegraphics[width=0.89\linewidth]{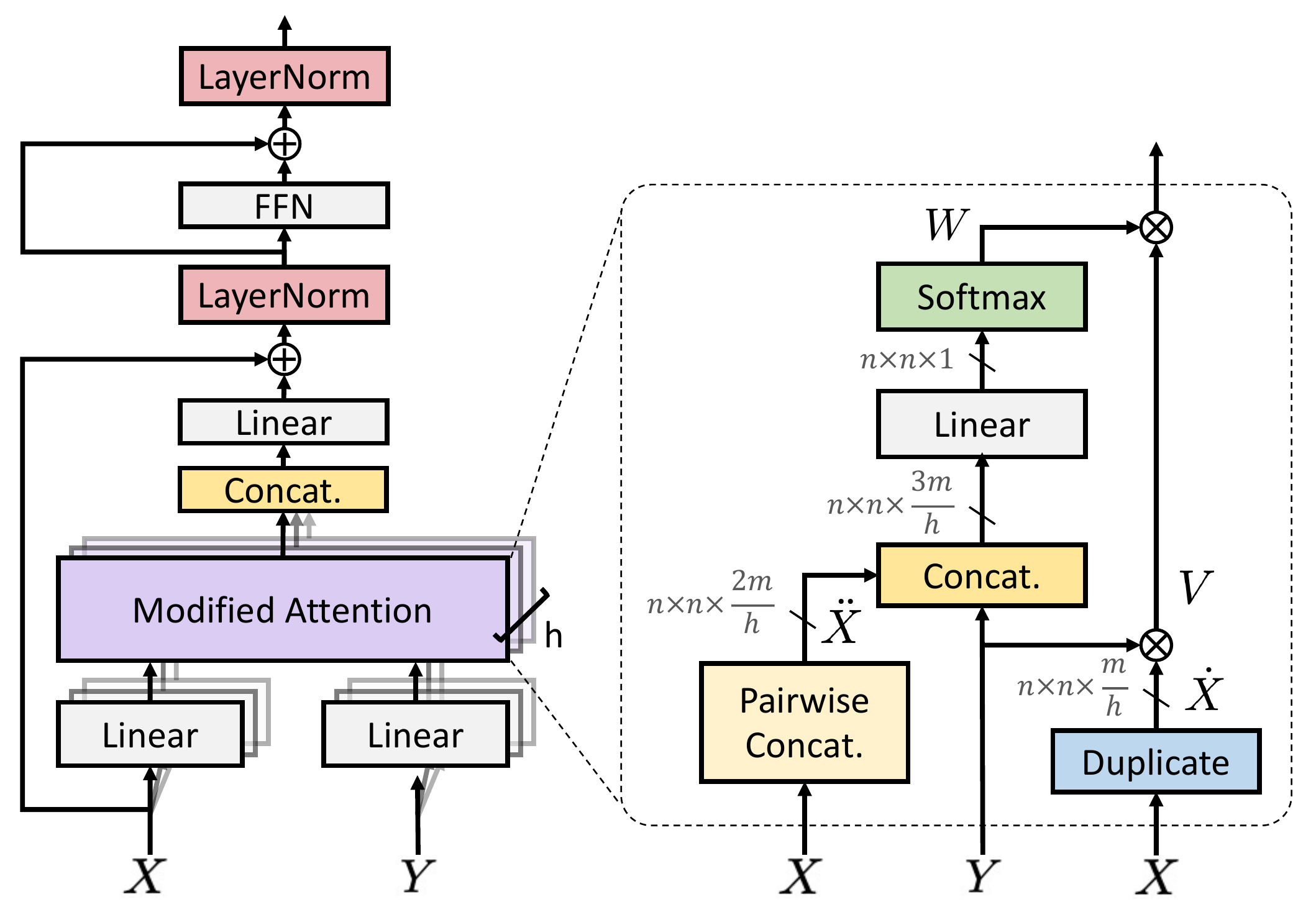}
    \caption{Architecture of the modified transformer encoder layer (left) and its attention module (right). FFN stands for feedforward network~\cite{xfmer}. ``Pairwise concat.'' refers to the operation of pairing up all tokens and concatenating the features. ``Duplicate'' refers to the operation of repeating the features along a new dimension.}
    \label{fig:modified-layer}
\end{figure}

More formally, given the detections returned by DETR, we first apply non-maximum suppression and thresholding. This leaves a smaller set $\{d_i\}_{i=1}^{n}$, where a detection $d_i=(\bb_i, s_i, c_i, \bx_i)$ consists of the box coordinates $\bb_i \in \reals^4$, the confidence score $s_i \in [0, 1]$, the object class $c_i \in \cK$ for a set of object categories $\cK$, and the object query or feature $\bx_i \in \reals^{m}$. We compute the pairwise box positional encodings $\{\by_{i, j} \in \reals^m\}_{i, j=1}^{n}$ as outlined above. 
We denote the collection of unary tokens by $X \in \reals^{n \times m}$ and the pairwise positional encodings by $Y \in \reals^{n \times n \times m}$. The complete structure of the modified transformer encoder layer is shown in \cref{fig:modified-layer}. For brevity of exposition, let us assume that the number of heads $h$ is 1, and define
\begin{align}
    \dot{X} \in \reals^{n \times n \times m},\: \dot{X}_i & \triangleq X \in \reals^{n \times m}, \\
    \ddot{X} \in \reals^{n \times n \times 2m},\: \ddot{\bx}_{i,j} & \triangleq \bx_{i} \oplus \bx_{j} \in \reals^{2m},
\end{align}
where $\oplus$ denotes vector concatenation. That is, the tensors $\dot{X}$ and $\ddot{X}$ are the results of duplication and pairwise concatenation. The equivalent values and attention weights can then be computed as
\begin{align}
    V &= \dot{X} \otimes Y, \\
    W &= \text{softmax}( (\ddot{X} \oplus Y) \bw + b ),
\end{align}
where $\otimes$ denotes elementwise product and $\bw \in \reals^{3m}$ and $b \in \reals$ are the parameters of the linear layer. The output of the attention layer is then computed as $W \otimes V$.
Additional details can be found in~\cref{app:me}.

\subsection{Competitive layer}

To compute the set of pairwise tokens, we form all pairs of distinct unary tokens and remove those where the first token is not human, as object--object pairs are beyond the scope of HOI detection. We denote the resulting set as $\{p_k = (\bx_i, \bx_j, \by_{i, j}) \mid i \neq j, c_i = ``\text{human}"\}$. We then compute the pairwise tokens from the unary tokens and positional encodings via multi-branch fusion (MBF)~\cite{scg} as
\begin{equation}
    \bz_k = \text{MBF}(\bx_i \oplus \bx_j, \by_{i, j}).
\end{equation}
Specifically, the MBF module fuses two modalities in multiple homogeneous branches and return a unified feature representation. For completeness, full details are provided in~\cref{app:mbf}. Last, the set of pairwise tokens are fed into an additional transformer encoder layer, allowing the network to compare the HOI candidates, before an MLP predicts each HOI pair's action classification logits $\widetilde{\bs}$.

\subsection{Training and inference}

To make full use of the pre-trained object detector, we incorporate the object confidence scores into the final scores of each human--object pair. Denoting the action logits of the $k^{\text{th}}$ pair $p_k$ as $\widetilde{\bs}_k$, the final scores are computed as
\begin{equation}
    \bs_k=(s_i)^\lambda \cdot (s_j)^\lambda \cdot \sigma(\widetilde{\bs}_k),
    \label{eq:scores}
\end{equation}
where $\lambda > 1$ is a constant used during inference to suppress overconfident objects~\cite{scg} and $\sigma$ is the sigmoid function. We use focal loss\footnote{Final scores in \cref{eq:scores} are normalised to the interval $[0, 1]$. In training, we instead recover the scale prior to normalisation and use the corresponding loss with logits for numerical stability. See more details in~\cref{app:loss}.}~\cite{retinanet} for action classification to counter the imbalance between positive and negative examples. Following previous practice~\cite{no-frills,scg}, we only compute the loss on valid action classes for each object type, specified by the dataset. During inference, scores for invalid combinations of actions and objects (\eg, \textit{eating a car}) are zeroed out.

\section{Experiments}

\begin{table*}[t]\small
    \centering
    \caption{Comparison of HOI detection performance (mAP$\times100$) on the HICO-DET~\cite{hicodet} and V-COCO~\cite{vcoco} test sets. The highest result in each section is highlighted in bold.}
    \label{tab:results}
    \begin{tabularx}{\linewidth}{@{\extracolsep{\fill}} l l cccccccc}
        \toprule
    & & \multicolumn{6}{c}{\textbf{HICO-DET}} & \multicolumn{2}{c}{\textbf{V-COCO}} \\ [4pt]
    & & \multicolumn{3}{c}{Default Setting} & \multicolumn{3}{c}{Known Objects Setting} & & \\ 
    \cline{3-5}\cline{6-8}\cline{9-10} \\ [-8pt]
        \textbf{Method} & \textbf{Backbone} & Full & Rare & Non-rare & Full & Rare & Non-rare & AP$_{role}^{S1}$ & AP$_{role}^{S2}$ \\
        \midrule
        HO-RCNN~\cite{hicodet} & CaffeNet & 7.81 & 5.37 & 8.54 & 10.41 & 8.94 & 10.85 & - & - \\
        InteractNet~\cite{interactnet} & ResNet-50-FPN & 9.94 & 7.16 & 10.77 & - & - & - & 40.0 & - \\
        GPNN~\cite{gpnn} & ResNet-101 & 13.11 & 9.34 & 14.23 & - & - & - & 44.0 & - \\
        TIN~\cite{tin} & ResNet-50 & 17.03 & 13.42 & 18.11 & 19.17 & 15.51 & 20.26 & 47.8 & 54.2 \\
        Gupta \etal~\cite{no-frills} & ResNet-152 & 17.18 & 12.17 & 18.68 & - & - & - & - & - \\
        VSGNet~\cite{vsgnet} & ResNet-152 & 19.80 & 16.05 & 20.91 & - & - & - & 51.8 & 57.0 \\
    DJ-RN~\cite{djrn} & ResNet-50 & 21.34 & 18.53 & 22.18 & 23.69 & 20.64 & 24.60 & - & - \\
    PPDM~\cite{ppdm} & Hourglass-104 & 21.94 & 13.97 & 24.32 & 24.81 & 17.09 & 27.12 & - & - \\
    VCL~\cite{vcl} & ResNet-50 & 23.63 & 17.21 & 25.55 & 25.98 & 19.12 & 28.03 & 48.3 & - \\
    ATL~\cite{atl} & ResNet-50 & 23.81 & 17.43 & 27.42 & 27.38 & 22.09 & 28.96 & - & - \\
    DRG~\cite{drg} & ResNet-50-FPN & 24.53 & 19.47 & 26.04 & 27.98 & 23.11 & 29.43 & 51.0 & - \\
    IDN~\cite{idn} & ResNet-50 & 24.58 & 20.33 & 25.86 & 27.89 & 23.64 & 29.16 & 53.3 & 60.3 \\
    HOTR~\cite{hotr} & ResNet-50 & 25.10 & 17.34 & 27.42 & - & - & - & 55.2 & \textbf{64.4} \\
    FCL~\cite{fcl} & ResNet-50 & 25.27 & 20.57 & 26.67 & 27.71 & 22.34 & 28.93 & 52.4 & - \\
    HOI-Trans~\cite{hoitrans} & ResNet-101 & 26.61 & 19.15 & 28.84 & 29.13 & 20.98 & 31.57 & 52.9 & - \\
    AS-Net~\cite{asnet} & ResNet-50 & 28.87 & 24.25 & 30.25 & 31.74 & 27.07 & 33.14 & 53.9 & - \\
        SCG~\cite{scg} & ResNet-50-FPN & 29.26 & \textbf{24.61} & 30.65 & \textbf{32.87} & \textbf{27.89} & \textbf{34.35} & 54.2 & 60.9 \\
    QPIC~\cite{qpic} & ResNet-101 & \textbf{29.90} & 23.92 & \textbf{31.69} & 32.38 & 26.06 & 34.27 & \textbf{58.8} & 61.0 \\
    \midrule
    Ours (UPT) & ResNet-50 & 31.66 & 25.94 &  33.36 & 35.05 & 29.27 & 36.77 & 59.0 & 64.5 \\
    Ours (UPT) & ResNet-101 & 32.31 & 28.55 & 33.44 & 35.65 & \textbf{31.60} & 36.86 & 60.7 & 66.2 \\
    Ours (UPT) & ResNet-101-DC5 & \textbf{32.62} & \textbf{28.62} & \textbf{33.81} & \textbf{36.08} & 31.41 & \textbf{37.47} & \textbf{61.3} & \textbf{67.1} \\
        \bottomrule
    \end{tabularx}
\end{table*}

In this section, we first demonstrate that the proposed unary--pairwise transformer achieves state-of-the-art performance on both the HICO-DET~\cite{hicodet} and V-COCO~\cite{vcoco} datasets, outperforming the next best method by a significant margin. We then provide a thorough analysis on the effects of the cooperative and competitive layers. In particular, we show that the cooperative layer increases the scores of positive examples while the competitive layer suppresses those of the negative examples. We then visualise the attention weights for specific images, and show how these behaviours are achieved by the attention mechanism.
At inference time, our method with ResNet50~\cite{resnet} runs at 24 FPS on a single GeForce RTX 3090 device.

\paragraph{Datasets:}
HICO-DET~\cite{hicodet} is a large-scale HOI detection dataset with $37\,633$ training images, $9\,546$ test images, $80$ object types, $117$ actions, and $600$ interaction types. The dataset has $117\,871$ human--object pairs with annotated bounding boxes in the training set and $33\,405$ in the test set.
V-COCO~\cite{vcoco} is much smaller in scale, with $2\,533$ training images, $2\,867$ validation images, $4\,946$ test images, and only $24$ different actions.

\subsection{Implementation details}

We fine-tune the DETR model on the HICO-DET and V-COCO datasets prior to training and then freeze its weights. For HICO-DET, we use the publicly accessible DETR models pre-trained on MS COCO~\cite{coco}. However, for V-COCO, as its test set is contained in the COCO val2017 subset, we first pre-train DETR models from scratch on MS COCO, excluding those images in the V-COCO test set. For the interaction head, we filter out detections with scores lower than $0.2$, and sample at least $3$ and up to $15$ humans and objects each, prioritising high scoring ones. For the hidden dimension of the transformer, we use $m=256$, the same as DETR. Additionally, we set $\lambda$ to $1$ during training and $2.8$ during inference~\cite{scg}. For the hyperparameters used in the focal loss, we use the same values as SCG~\cite{scg}.

We apply a few data augmentation techniques used in other detectors~\cite{detr,qpic}. Inputs images are scaled such that the shortest side is at least $480$ and at most $800$ pixels. The longest side is limited at $1333$ pixels. Additionally, each image is cropped with a probability of $0.5$ to a random rectangle with each side being at least $384$ pixels and at most $600$ pixels before being scaled. We also apply colour jittering, where the brightness, contrast and saturation values are adjusted by a random factor between $0.6$ to $1.4$. We use AdamW~\cite{adamw} as the optimiser with an initial learning rate of $10^{-4}$. All models are trained for $20$ epochs with a learning rate reduction at the $10^{\text{th}}$ epoch by a factor of $10$. Training is conducted on $8$ GeForce GTX TITAN X devices, with a batch size of $2$ per GPU---an effective batch size of $16$.

\subsection{Comparison with state-of-the-art methods}

\begin{table*}[t]\small
    \caption{Comparing the effect of the cooperative (coop.) and competitive (comp.) layers on the interaction scores. We report the change in the interaction scores as the layer in the $\Delta$ Architecture column is added to the reference network, for positives, easy negatives and hard negatives, with the number of examples in parentheses. As indicated by the bold numbers, the cooperative layer significantly increases the scores of positive examples while the competitive layer suppresses hard negative examples. Together, these layers widen the gap between scores of positive and negative examples, improving the detection mAP.}
    \label{tab:delta}
    \setlength{\tabcolsep}{3pt} 
    \begin{tabularx}{\linewidth}{@{\extracolsep{\fill}} l l c c c c c c}
        \toprule
        & & \multicolumn{2}{c}{$\Delta$ \textbf{Positives} ($25\,391$)} & \multicolumn{2}{c}{$\Delta$ \textbf{Easy Negatives} ($3\,903\,416$)} & \multicolumn{2}{c}{$\Delta$ \textbf{Hard Negatives} ($510\,991$)}\\ [4pt]
        \cline{3-4} \cline{5-6} \cline{7-8} \\ [-8pt]
        \textbf{Reference} & $\Delta$ \textbf{Architecture} & Mean & Median & Mean & Median & Mean & Median \\
        \midrule
        Ours w/o coop. layer & + coop. layer & \textbf{+0.1487} & +0.1078 & +0.0001 & +0.0000 & +0.0071 & +0.0000 \\
        Ours w/o comp. layer & + comp. layer & -0.0463 & -0.0310 & -0.0096 & -0.0024 & \textbf{-0.1080} & -0.0922 \\
        Ours w/o both layers & + both layers & \textbf{+0.0799} & +0.0390 & -0.0076 & -0.0018 & \textbf{-0.0814} & -0.0748 \\
        \bottomrule
    \end{tabularx}
\end{table*}

\begin{figure*}[t]
    \begin{subfigure}[t]{0.33\linewidth}
        \centering
        \includegraphics[width=\linewidth]{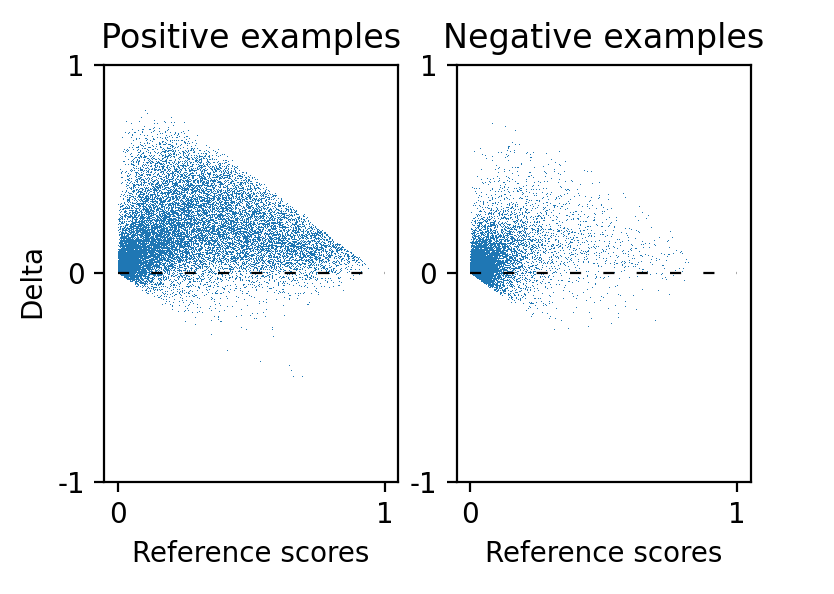}
        \caption{\cref{tab:delta} first row}
        \label{fig:scatter-left}
    \end{subfigure}
    \begin{subfigure}[t]{0.33\linewidth}
        \centering
        \includegraphics[width=\linewidth]{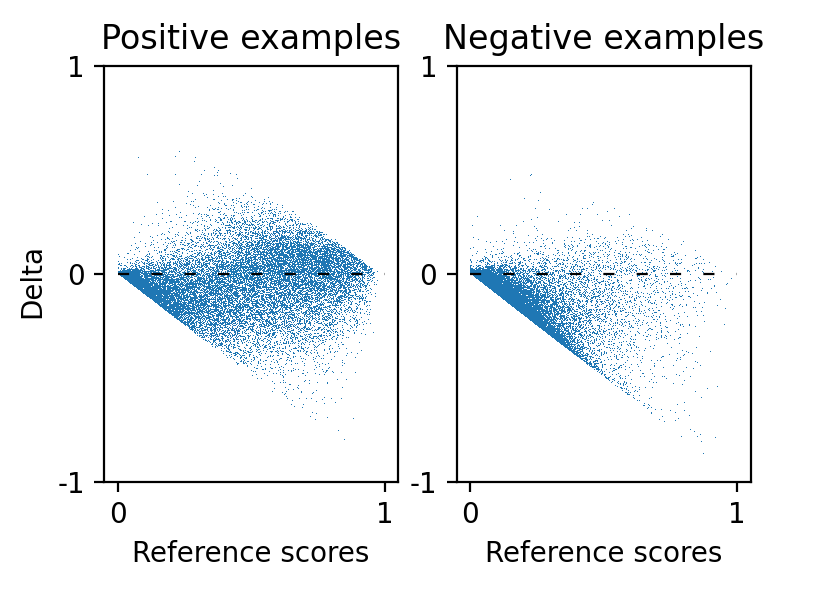}
        \caption{\cref{tab:delta} second row}
        \label{fig:scatter-mid}
    \end{subfigure}
    \begin{subfigure}[t]{0.33\linewidth}
        \centering
        \includegraphics[width=\linewidth]{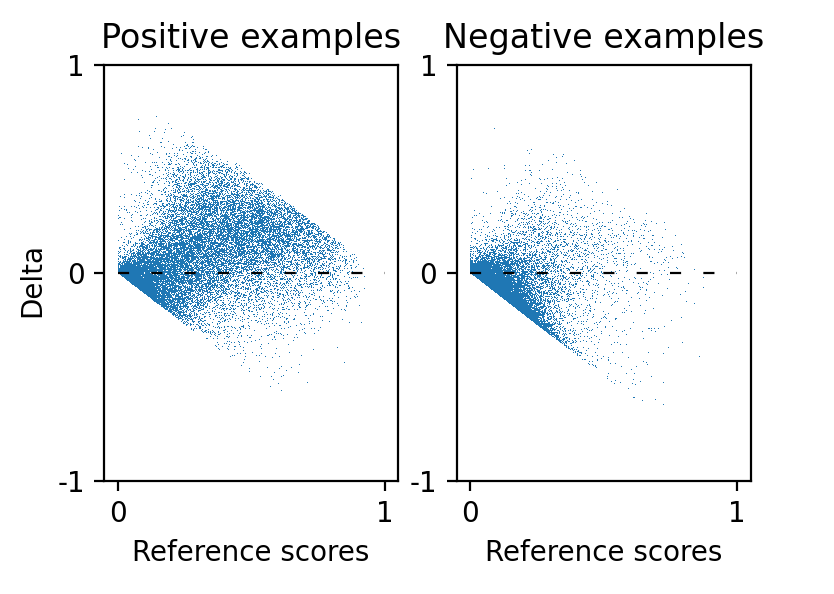}
        \caption{\cref{tab:delta} third row}
        \label{fig:scatter-right}
    \end{subfigure}
    \caption{Change in the interaction score (delta) with respect to the reference score. \subref{fig:scatter-left} The distribution of score deltas when adding the cooperative layer (first row of \cref{tab:delta}). \subref{fig:scatter-mid} Adding the competitive layer to the model (second row). \subref{fig:scatter-right} Adding both layers (last row). For visualisation purposes, only $20\%$ of the negatives are sampled and displayed.
    }
    \label{fig:scatter}
\end{figure*}

The performance of our model is compared to existing methods on the HICO-DET~\cite{hicodet} and V-COCO~\cite{vcoco} datasets in \cref{tab:results}. There are two different settings for evaluation on HICO-DET. \textit{Default Setting}: A detected human--object pair is considered matched with a ground truth pair, if the minimum intersection over union (IoU) between the human boxes and object boxes exceeds $0.5$. Amongst all matched pairs, the one with the highest score is considered the true positive while others are false positives. Pairs without a matched ground truth are also considered false positives. \textit{Known Objects Setting}: Besides the aforementioned criteria, this setting assumes the set of object types in ground truth pairs are known. Therefore, detected pairs with an object type outside the set are removed automatically, thus reducing the difficulty of the problem. For V-COCO, the average precision (AP) is computed under two scenarios, differentiated by the superscripts $S1$ and $S2$. This is to account for missing objects due to occlusion. For scenario $1$, empty object boxes should be predicted in case of occlusion for a detected pair to be considered a match with the corresponding ground truth, while for scenario $2$, object boxes are always assumed to be matched in such cases.

We report our model's performance for three different backbone networks. Notably, our model with the lightest-weight backbone already outperforms the next best method by a significant margin in almost every category. This gap is further widened with more powerful backbone networks. In particular, since the backbone CNN and object detection transformer are detached from the computational graph, our model has a small memory footprint. This allows us to use a higher-resolution feature map by removing the stride in the $5^{\text{th}}$ convolutional block (C5) of ResNet~\cite{resnet}, which has been shown to improve detection performance on small objects~\cite{detr}. We denote this as dilated C5 (DC5).

\subsection{Macroscopic effects of the interaction head}
\label{sec:macro}

In this section, we compare the effects of the unary (cooperative) and pairwise (competitive) layers on the HICO-DET test set, with ResNet50~\cite{resnet} as the CNN backbone.
Since the parameters in the object detector are kept frozen for our model, the set of detections processed by the downstream network remains the same, regardless of any architectural changes in the interaction head. This allows us to compare how different variants of our model perform on the same human--object pairs. To this end, we collected the predicted interaction scores for all human--object pairs over the test set and compare how adding certain layers influence them. In \cref{tab:delta}, we show some statistics on the change of scores upon an architectural modification. In particular, note that the vast majority of collected pairs are easy negatives with scores close to zero. For analysis, we divide the negative examples into easy and hard, where we define an easy negative as one with a score lower than $0.05$ as predicted by the ``Ours w/o both layers'' model, which accounts for $90\%$ of the negative examples. In addition, we also show the distribution of the change in score with respect to the reference score as scatter plots in \cref{fig:scatter}. The points are naturally bounded by the half-spaces $0 \leq x+y \leq 1$.

Notably, adding the cooperative layer results in a significant average increase ($+0.15$) in the scores of positive examples, with little effect on the negative examples. This can be seen in \cref{fig:scatter-left} as well, where the score changes for almost all positive examples are larger than zero.
In contrast, adding the competitive layer leads to a significant average decrease ($-0.11$) in the scores of hard negative examples, albeit with a small decrease in the score of positive examples as well. This minor decrease is compensated by the cooperative layer as shown in the last row of \cref{tab:delta}. Furthermore, looking at \cref{fig:scatter-mid}, we can see a dense mass near the line $y=-x$, which indicates that many negative examples have had their scores suppressed to zero.

\begin{table}[t]\small
    \caption{Effect of the cooperative and competitive layers on the HICO-DET test set under the default settings.}
    \label{tab:ablation}
    \setlength{\tabcolsep}{3pt} 
    \begin{tabularx}{\linewidth}{l C C C}
        \toprule
        \textbf{Model} & \textbf{Full} & \textbf{Rare} & \textbf{Non-rare} \\
        \midrule
        Ours w/o both layers & 29.22 & 23.09 & 31.05 \\
        Ours w/o comp. layer & 30.78 & 24.92 & 32.53 \\
        Ours w/o coop. layer & 30.68 & 24.69 & 32.47 \\
        Ours w/o pairwise pos. enc. & 29.98 & 23.72 &  31.64 \\
        \midrule
        Ours ($1 \times$ coop., $1 \times$ comp.) & 31.33 & 26.02 & 32.91 \\
        Ours ($1 \times$ coop., $2 \times$ comp.) & 31.62 & \textbf{26.18} & 33.24 \\
        Ours ($2 \times$ coop., $1 \times$ comp.) & \textbf{31.66} & 25.94 & \textbf{33.36} \\
        \bottomrule
    \end{tabularx}
\end{table}

\paragraph{Ablation study:}
In \cref{tab:ablation}, we ablate the effect of different design decisions on performance. Adding the cooperative and competitive layers individually improves the performance by around $1.5$~mAP, while adding both layers jointly improves by over $2$~mAP. We also demonstrate the significance of the pairwise position encodings by removing them from the modified encoder and the multi-branch fusion module. This results in a 1.3~mAP decrease. Finally, we observe a slight improvement (0.3~mAP) when adding an additional cooperative or competitive layer, but no further improvements with more layers. As the competitive layer is more costly, we use two cooperative layers.

\subsection{Microscopic effects of the interaction head}
\label{sec:micro}

\begin{figure}[t]
    \centering
    \includegraphics[height=0.36\linewidth]{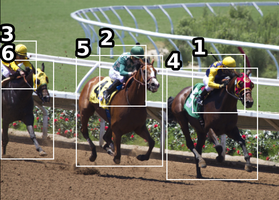} \hspace{3pt}
    \includegraphics[height=0.36\linewidth]{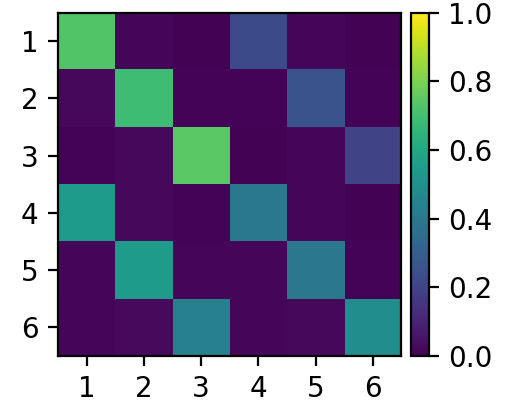}
    \caption{Detected human and object instances (left) and the unary attention map for these instances (right).}
    \label{fig:unary_attn}
\end{figure}

\begin{figure}[t]
    \centering
    \includegraphics[width=\linewidth]{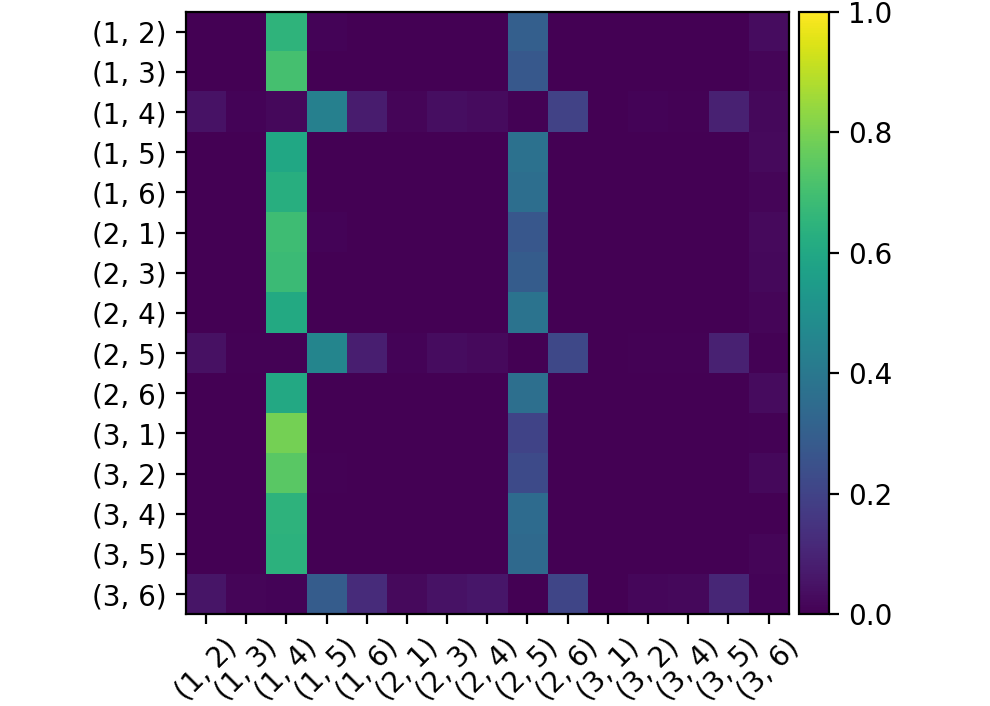}
    \caption{Pairwise attention map for the human and object instances in \cref{fig:unary_attn}.}
    \label{fig:pairwise_attn}
\end{figure}

\begin{figure*}[t]
    
    \begin{subfigure}[t]{0.19\linewidth}
        \centering
        \includegraphics[height=0.68\linewidth]{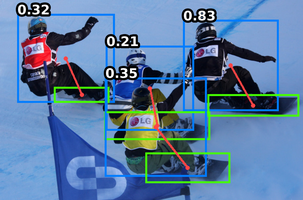}
        \caption{\textit{standing on a snowboard}}
        \label{fig:standing-on-snowboard}
    \end{subfigure}
    \hfill%
    \begin{subfigure}[t]{0.19\linewidth}
        \centering
        \includegraphics[height=0.68\linewidth]{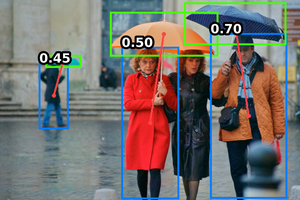}
        \caption{\textit{holding an umbrella}}
        \label{fig:holding-umbrella}
    \end{subfigure}
    \hfill%
    \begin{subfigure}[t]{0.19\linewidth}
        \centering
        \includegraphics[height=0.68\linewidth]{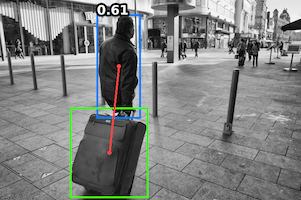}
        \caption{\textit{carrying a suitcase}}
        \label{fig:carrying-suitcase}
    \end{subfigure}
    \hfill%
    \begin{subfigure}[t]{0.19\linewidth}
        \centering
        \includegraphics[height=0.68\linewidth]{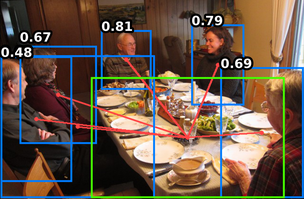}
        \caption{\textit{sitting at a dining table}}
        \label{fig:sitting-at-dinning-table}
    \end{subfigure}
    \hfill%
    \begin{subfigure}[t]{0.19\linewidth}
        \centering
        \includegraphics[height=0.68\linewidth]{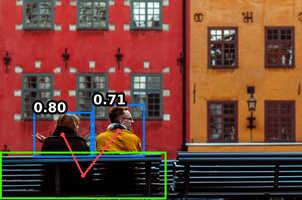}
        \caption{\textit{sitting on a bench}}
        \label{fig:sitting-on-bench}
    \end{subfigure}
    
    \begin{subfigure}[t]{0.19\linewidth}
        \centering
        \includegraphics[height=0.68\linewidth]{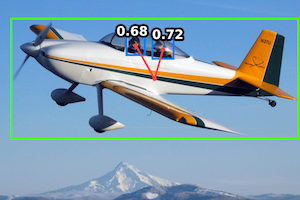}
        \caption{\textit{flying an airplane}}
        \label{fig:flying-airplane}
    \end{subfigure}
    \hfill%
    \begin{subfigure}[t]{0.19\linewidth}
        \centering
        \includegraphics[height=0.68\linewidth]{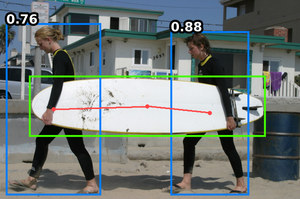}
        \caption{\textit{holding a surfboard}}
        \label{fig:holding-surfboard}
    \end{subfigure}
    \hfill%
    \begin{subfigure}[t]{0.19\linewidth}
        \centering
        \includegraphics[height=0.68\linewidth]{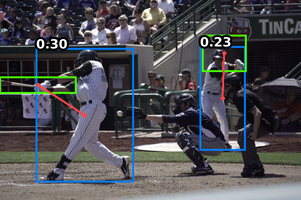}
        \caption{\textit{wielding a baseball bat}}
        \label{fig:wielding-baseball-bat}
    \end{subfigure}
    \hfill%
    \begin{subfigure}[t]{0.19\linewidth}
        \centering
        \includegraphics[height=0.68\linewidth]{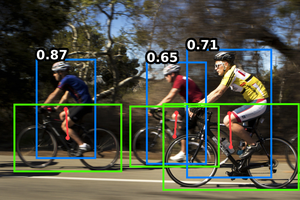}
        \caption{\textit{riding a bike}}
        \label{fig:riding-bike}
    \end{subfigure}
    \hfill%
    \begin{subfigure}[t]{0.19\linewidth}
        \centering
        \includegraphics[height=0.68\linewidth]{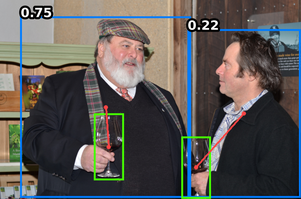}
        \caption{\textit{holding a wineglass}}
        \label{fig:holding-wineglass}
    \end{subfigure}
    
    \caption{Qualitative results of detected HOIs. Interactive human--object pairs are connected by red lines, with the interaction scores overlaid above the human box. Pairs with scores lower than $0.2$ are filtered out.}
    \label{fig:qualitative}
\end{figure*}

\begin{figure*}[t]
    \centering
    
    \begin{subfigure}[t]{0.19\linewidth}
    \centering
    \includegraphics[height=0.67\linewidth]{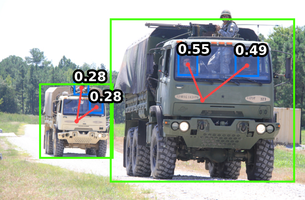}
    \caption{\textit{driving a truck}}
    \label{fig:driving-truck}
    \end{subfigure} \hfill
\begin{subfigure}[t]{0.19\linewidth}
    \includegraphics[height=0.67\linewidth]{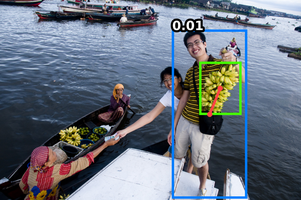}
    \caption{\textit{buying bananas}}
    \label{fig:buying-bananas}
\end{subfigure} \hfill
    \begin{subfigure}[t]{0.19\linewidth}
    \centering
    \includegraphics[height=0.67\linewidth]{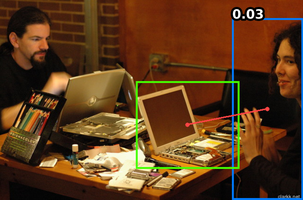}
    \caption{\textit{repairing a laptop}}
    \label{fig:repairing-laptop}
    \end{subfigure} \hfill
    \begin{subfigure}[t]{0.19\linewidth}
    \centering
    \includegraphics[height=0.67\linewidth]{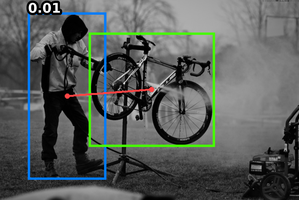}
    \caption{\textit{washing a bicycle}}
    \label{fig:washing-bike}
    \end{subfigure} \hfill
    \begin{subfigure}[t]{0.19\linewidth}
    \centering
    \includegraphics[height=0.67\linewidth]{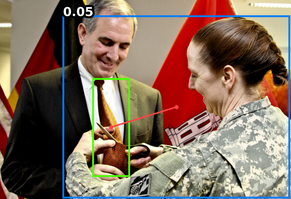}
    \caption{\textit{cutting a tie}}
    \label{fig:cutting-tie}
    \end{subfigure}

    \caption{Failure cases often occur when there is ambiguity in the interaction~\subref{fig:driving-truck},~\subref{fig:buying-bananas},~\subref{fig:repairing-laptop} or a lack of training data~\subref{fig:repairing-laptop},~\subref{fig:washing-bike},~\subref{fig:cutting-tie}.}
    \label{fig:failure}
\end{figure*}

In this section, we focus on a specific image and visualise the effect of attention in our cooperative and competitive layers. In \cref{fig:unary_attn}, we display a detection-annotated image and its associated attention map from the unary (cooperative) layer. The human--object pairs $(1, 4)$, $(2, 5)$ and $(3, 6)$ are engaged in the interaction \textit{riding a horse}. Excluding attention weights along the diagonal, we see that the corresponding human and horse instances attend to each other.
We hypothesise that attention between pairs of unary tokens (e.g., $1$ and $4$) helps increase the interaction scores for the corresponding pairs. To validate this hypothesis, we manually set the attention logits between the three positive pairs to minus infinity, thus zeroing out the corresponding attention weights. The effect of this was an average decrease of $0.06$ ($8\%$) in the interaction scores for the three pairs, supporting the hypothesis.

In \cref{fig:pairwise_attn}, we visualise the attention map of the pairwise (competitive) layer. Notably, all human--object pairs attend to the interactive pairs $(1, 4)$, $(2, 5)$ and $(3, 6)$ in decreasing order, except for the interactive pairs themselves. We hypothesise that attention is acting here to have the dominant pairs suppress the other pairs. To investigate, we manually set the weights such that the three interactive pairs all attend to $(1, 4)$ as well, with a weight of $1$. This resulted in a decrease of their interaction scores by $0.08$ ($11\%$). We then instead zeroed out the attention weights between the rest of the pairs and ($1, 4$), which resulted in a small increase in the scores of negative pairs. These results together suggest that attention in the competitive layer is acting as a soft version of non-maximum suppression, where pairs less likely to foster interactions attend to, and are suppressed by, the most dominant pairs. See~\cref{app:qual} for more examples.

\subsection{Qualitative results and limitations}
\vspace{-8pt}

In \cref{fig:qualitative}, we present several qualitative examples of successful HOI detections, where our model accurately localises the human and object instances and assigns high scores to the interactive pairs. For example, in \cref{fig:holding-umbrella}, our model correctly identifies the subject of an interaction (the lady in red) despite her proximity to a non-interactive human (the lady in black). We also observe in \cref{fig:standing-on-snowboard} that our model becomes less confident when there is overlap and occlusion. This stems from the use of object detection scores in our model. Confusion in the object detector often translates to confusion in action classification.
We also show five representative failure cases for our model, illustrating its limitations. In \cref{fig:driving-truck}, due to the indefinite position of drivers in the training set (and real life), the model struggled to identify the driver. For \cref{fig:washing-bike}, the model failed to recognise the interaction due to a lack of training data ($1$ training example), even though the action is well-defined. Overall, ambiguity in the actions and insufficient data are the biggest challenges for our model.
Another limitation, specific to our model, is that the computation and memory requirements of our pairwise layer scale quadratically with the number of unary tokens. For scenes involving many interactive humans and objects, this becomes quite costly.
Moreover, since the datasets we used are limited, we may expect poorer performance on 
    data in the wild, where image resolution, lighting condition, etc. may be less controlled.

\section{Conclusion}

In this paper, we have proposed a two-stage detector of human--object interactions using a novel transformer architecture that exploits both unary and pairwise representations of the human and object instances. Our model not only outperforms the current state-of-the-art---a one-stage detector---but also consumes much less time and memory to train. Through extensive analysis, we demonstrate that attention between unary tokens acts to increase the scores of positive examples, while attention between pairwise tokens acts like non-maximum suppression, reducing the scores of negative examples. We show that these two effects are complementary, and together boost performance significantly.

\vspace{-10pt}
\paragraph{Potential negative societal impact:}
Transformer models are large and computationally-expensive, and so have a significant negative environmental impact. To mitigate this, we use pre-trained models and a two-stage architecture, since fine-tuning an existing model requires less resources, as does training a single stage with the other stage fixed. There is also the potential for HOI detection models to be misused, such as for unauthorised surveillance, which disproportionately affects minority and marginalised communities.

\vspace{-10pt}
\paragraph{Acknowledgments:}
We are grateful for support from Continental AG (D.C.). We would also like to thank Jia-Bin Huang and Yuliang Zou for their help with the reproduction of some experiment results.

\clearpage

{\small
\bibliographystyle{ieee_fullname}
\bibliography{egbib}

\begin{thebibliography}{10}\itemsep=-1pt

\bibitem{detr}
Nicolas Carion, Francisco Massa, Gabriel Synnaeve, Nicolas Usunier, Alexander
  Kirillov, and Sergey Zagoruyko.
\newblock End-to-end object detection with transformers.
\newblock In {\em Eur. Conf. Comput. Vis.}, 2020.

\bibitem{hicodet}
Yu-Wei Chao, Yunfan Liu, Xieyang Liu, Huayi Zeng, and Jia Deng.
\newblock Learning to detect human-object interactions.
\newblock In {\em Proceedings of the IEEE Winter Conference on Applications of
  Computer Vision}, 2018.

\bibitem{asnet}
Mingfei Chen, Yue Liao, Si Liu, Zhiyuan Chen, Fei Wang, and Chen Qian.
\newblock Reformulating hoi detection as adaptive set prediction.
\newblock In {\em IEEE Conf. Comput. Vis. Pattern Recog.}, 2021.

\bibitem{vit}
Alexey Dosovitskiy, Lucas Beyer, Alexander Kolesnikov, Dirk Weissenborn,
  Xiaohua Zhai, Thomas Unterthiner, Mostafa Dehghani, Matthias Minderer, Georg
  Heigold, Sylvain Gelly, Jakob Uszkoreit, and Neil Houlsby.
\newblock An image is worth 16x16 words: Transformers for image recognition at
  scale.
\newblock In {\em Int. Conf. Learn. Represent.}, 2021.

\bibitem{drg}
Chen Gao, Jiarui Xu, Yuliang Zou, and Jia-Bin Huang.
\newblock {DRG}: Dual relation graph for human-object interaction detection.
\newblock In {\em Eur. Conf. Comput. Vis.}, 2020.

\bibitem{interactnet}
Georgia Gkioxari, Ross Girshick, Piotr Doll\'{a}r, and Kaiming He.
\newblock Detecting and recognizing human-object interactions.
\newblock In {\em IEEE Conf. Comput. Vis. Pattern Recog.}, 2018.

\bibitem{vcoco}
Saurabh Gupta and Jitendra Malik.
\newblock Visual semantic role labeling.
\newblock {\em arXiv preprint arXiv:1505.04474}, 2015.

\bibitem{no-frills}
Tanmay Gupta, Alexander Schwing, and Derek Hoiem.
\newblock No-frills human-object interaction detection: Factorization, layout
  encodings, and training techniques.
\newblock In {\em Int. Conf. Comput. Vis.}, 2019.

\bibitem{resnet}
Kaiming He, Xiangyu Zhang, Shaoqing Ren, and Jian Sun.
\newblock Deep residual learning for image recognition.
\newblock In {\em IEEE Conf. Comput. Vis. Pattern Recog.}, pages 770--778,
  2016.

\bibitem{vcl}
Zhi Hou, Xiaojiang Peng, Yu Qiao, and Dacheng Tao.
\newblock Visual compositional learning for human-object interaction detection.
\newblock In {\em Eur. Conf. Comput. Vis.}, 2020.

\bibitem{atl}
Zhi Hou, Baosheng Yu, Yu Qiao, Xiaojiang Peng, and Dacheng Tao.
\newblock Affordance transfer learning for human-object interaction detection.
\newblock In {\em IEEE Conf. Comput. Vis. Pattern Recog.}, 2021.

\bibitem{fcl}
Zhi Hou, Baosheng Yu, Yu Qiao, Xiaojiang Peng, and Dacheng Tao.
\newblock Detecting human-object interaction via fabricated compositional
  learning.
\newblock In {\em IEEE Conf. Comput. Vis. Pattern Recog.}, 2021.

\bibitem{hotr}
Bumsoo Kim, Junhyun Lee, Jaewoo Kang, Eun-Sol Kim, and Hyunwoo~J. Kim.
\newblock Hotr: End-to-end human-object interaction detection with
  transformers.
\newblock In {\em IEEE Conf. Comput. Vis. Pattern Recog.}, 2021.

\bibitem{djrn}
Yong-Lu Li, Xinpeng Liu, Han Lu, Shiyi Wang, Junqi Liu, Jiefeng Li, and Cewu
  Lu.
\newblock Detailed 2d-3d joint representation for human-object interaction.
\newblock In {\em IEEE Conf. Comput. Vis. Pattern Recog.}, 2020.

\bibitem{idn}
Yong-Lu Li, Xinpeng Liu, Xiaoqian Wu, Yizhuo Li, and Cewu Lu.
\newblock Hoi analysis: Integrating and decomposing human-object interaction.
\newblock In {\em Adv. Neural Inform. Process. Syst.}, 2020.

\bibitem{tin}
Yong-Lu Li, Siyuan Zhou, Xijie Huang, Liang Xu, Ze Ma, Hao-Shu Fang, Yanfeng
  Wang, and Cewu Lu.
\newblock Transferable interactiveness knowledge for human-object interaction
  detection.
\newblock In {\em IEEE Conf. Comput. Vis. Pattern Recog.}, 2019.

\bibitem{ppdm}
Yue Liao, Si Liu, Fei Wang, Yanjie Chen, Chen Qian, and Jiashi Feng.
\newblock {PPDM}: Parallel point detection and matching for real-time
  human-object interaction detection.
\newblock In {\em IEEE Conf. Comput. Vis. Pattern Recog.}, 2020.

\bibitem{retinanet}
Tsung{-}Yi Lin, Priya Goyal, Ross~B. Girshick, Kaiming He, and Piotr
  Doll{\'{a}}r.
\newblock Focal loss for dense object detection.
\newblock In {\em Int. Conf. Comput. Vis.}, 2017.

\bibitem{coco}
Tsung-Yi Lin, Michael Maire, Serge Belongie, James Hays, Pietro Perona, Deva
  Ramanan, Piotr Dollár, and C.~Lawrence Zitnick.
\newblock Microsoft {COCO}: Common objects in context.
\newblock In {\em Eur. Conf. Comput. Vis.}, 2014.

\bibitem{train-xfmer}
Liyuan Liu, Xiaodong Liu, Jianfeng Gao, Weizhu Chen, and Jiawei Han.
\newblock Understanding the difficulty of training transformers.
\newblock In {\em Proceedings of the 2020 Conference on Empirical Methods in
  Natural Language Processing}, 2020.

\bibitem{swint}
Ze Liu, Yutong Lin, Yue Cao, Han Hu, Yixuan Wei, Zheng Zhang, Stephen Lin, and
  Baining Guo.
\newblock Swin transformer: Hierarchical vision transformer using shifted
  windows.
\newblock In {\em Int. Conf. Comput. Vis.}, pages 10012--10022, October 2021.

\bibitem{adamw}
Ilya Loshchilov and Frank Hutter.
\newblock Decoupled weight decay regularization.
\newblock In {\em Int. Conf. Learn. Represent.}, 2018.

\bibitem{gpnn}
Siyuan Qi, Wenguan Wang, Baoxiong Jia, Jianbing Shen, and Song-Chun Zhu.
\newblock Learning human-object interactions by graph parsing neural networks.
\newblock In {\em Eur. Conf. Comput. Vis.}, 2018.

\bibitem{fasterrcnn}
Shaoqing Ren, Kaiming He, Ross~B. Girshick, and Jian Sun.
\newblock Faster {R-CNN}: Towards real-time object detection with region
  proposal networks.
\newblock In {\em Adv. Neural Inform. Process. Syst.}, pages 91--99, 2015.

\bibitem{qpic}
Masato Tamura, Hiroki Ohashi, and Tomoaki Yoshinaga.
\newblock {QPIC}: Query-based pairwise human-object interaction detection with
  image-wide contextual information.
\newblock In {\em IEEE Conf. Comput. Vis. Pattern Recog.}, 2021.

\bibitem{vsgnet}
Oytun Ulutan, A~S~M Iftekhar, and B.~S. Manjunath.
\newblock {VSGNet}: Spatial attention network for detecting human object
  interactions using graph convolutions.
\newblock In {\em IEEE Conf. Comput. Vis. Pattern Recog.}, 2020.

\bibitem{xfmer}
Ashish Vaswani, Noam Shazeer, Niki Parmar, Jakob Uszkoreit, Llion Jones,
  Aidan~N Gomez, \L~ukasz Kaiser, and Illia Polosukhin.
\newblock Attention is all you need.
\newblock In {\em Adv. Neural Inform. Process. Syst.}, volume~30, 2017.

\bibitem{scg}
Frederic~Z. Zhang, Dylan Campbell, and Stephen Gould.
\newblock Spatially conditioned graphs for detecting human-object interactions.
\newblock In {\em Int. Conf. Comput. Vis.}, pages 13319--13327, October 2021.

\bibitem{hoitrans}
Cheng Zou, Bohan Wang, Yue Hu, Junqi Liu, Qian Wu, Yu Zhao, Boxun Li, Chenguang
  Zhang, Chi Zhang, Yichen Wei, and Jian Sun.
\newblock End-to-end human object interaction detection with hoi transformer.
\newblock In {\em IEEE Conf. Comput. Vis. Pattern Recog.}, 2021.

\end{thebibliography}
}

\appendix

\section{Pairwise positional encodings}
\label{app:pe}

We describe the details of the pairwise positional encodings as introduced in~\cref{sec:coop} of the main paper. Formally, denote a bounding box as $\bb = [x, y, w, h]^T \in [0, 1]^4$, where $x$ and $y$ represent the centre coordinates of the bounding box while $w$ and $h$ represent the width and height. Note that these values have been normalised by the image dimensions. For a pair of bounding boxes $\bb_1$ and $\bb_2$, we start by encoding the unary terms besides the box representation itself, including box areas and aspect ratios as below
\begin{equation}
    \bu = \bb_1 \oplus \bb_2 \oplus \left[w_1h_1, w_2h_2, \frac{w_1}{h_1}, \frac{w_2}{h_2} \right]^T,
    \label{eq:unary}
\end{equation}
where $\oplus$ denotes vector concatenation. We then proceed to encode the pairwise terms as follows
\begin{align}
    \label{eq:pairwise}
    \bp &= \left[\frac{w_1h_1}{w_2h_2}, \texttt{IoU}(\bb_1, \bb_2) \right]^T \oplus f(d_x) \oplus f(d_y), \\
    d_x &= \frac{x_1 - x_2}{w_1}, d_y = \frac{y_1 - y_2}{h_1}, \\
    f(d) &= \left[\texttt{ReLU}(d), \texttt{ReLU}(-d) \right]^T.
\end{align}
This includes additional features such as the ratio of box areas, intersection over union (IoU) and directional encodings that characterise the distance between box centres. Note that the directional encodings are normalised by the dimension of the first (human) bounding box instead of that of the image. In addition, function $f(\cdot)$ ensures the componentwise positivity of the feature vector. Finally, denote a multi-layer perceptron as MLP, the complete pairwise positional encoding is computed as below
\begin{equation}
    \by = \text{MLP}(\bu \oplus \bp \oplus \text{log}(\bu \oplus \bp + \epsilon)),
\end{equation}
where $\epsilon$ is a small constant added to the vector to avoid taking the logarithm of zero.

\section{Numerical stability in the loss function}
\label{app:loss}

For the sake of numerical stability, loss function for logits is often preferred to that for normalised scores. In our case, due to the fact that the final interaction score is the product of multiple factors, we cannot directly use the loss function for logits. Therefore, we first need to recover the scale prior to normalisation. Denote the normalised object confidence score and action logit as $\hat{y}_1 \in [0, 1]$ and $\hat{y}_2 \in \reals$ respectively, the final score is computed as $\hat{y} = \hat{y}_1 \cdot \sigma(\hat{y}_2)$, where $\sigma$ denotes the sigmoid function. We can then retrieve the corresponding logit $\tilde{y}$ as below
\begin{align}
    \tilde{y} &= \sigma^{-1}(\hat{y}), \\
    &= \text{log} \left( \frac{\hat{y}_1}{1 + \text{exp}(-\hat{y}_2) - \hat{y}_1} + \epsilon \right),
\end{align}
where $\epsilon$ is a small constant added to the term to avoid taking the logarithm of zero.

\section{Multi-branch fusion}
\label{app:mbf}

\begin{lstlisting}[
    float=*t,
    label={list:mbf},
    caption={PyTorch implementation of the multi-branch fusion module.}
]
import torch
import torch.nn as nn
import torch.nn.functional as F

class MultiBranchFusion(nn.Module):
    """
    Parameters:
    -----------
    appearance_size: int
        Size of the appearance features
    spatial_size: int
        Size of the spatial features
    hidden_state_size: int
        Size of the intermediate representations
    cardinality: int
        The number of homogeneous branches
    """
    def __init__(self,
        appearance_size: int = 256, spatial_size: int = 256,
        hidden_state_size: int = 256, cardinality: int = 8
    ) -> None:
        super().__init__()
        self.cardinality = cardinality
        sub_repr_size = int(hidden_state_size / cardinality)
        assert sub_repr_size * cardinality == hidden_state_size, \
            "The given representation size should be divisible by cardinality"

        self.fc_1 = nn.ModuleList([
            nn.Linear(appearance_size, sub_repr_size) for _ in range(cardinality)])
        self.fc_2 = nn.ModuleList([
            nn.Linear(spatial_size, sub_repr_size) for _ in range(cardinality)])
        self.fc_3 = nn.ModuleList([
            nn.Linear(sub_repr_size, hidden_state_size) for _ in range(cardinality)])

    def forward(self, appearance: torch.Tensor, spatial: torch.Tensor) -> torch.Tensor:
        return torch.stack([
            fc_3(F.relu(fc_1(appearance) * fc_2(spatial)))
            for fc_1, fc_2, fc_3 in zip(self.fc_1, self.fc_2, self.fc_3)
        ]).sum(dim=0)
\end{lstlisting}

The multi-branch fusion (MBF) module~\cite{scg} employs multiple homogeneous branches, wherein each branch maps the two input features into a subspace of reduced dimension and performs fusion (elementwise product by default). The resultant feature is then mapped back to the original size. Afterwards, elementwise sum is used to aggregate results across all branches. The reduced representation size in a branch is intentionally configured in a way that renders the total number of parameters independent of the number of branches. For brevity of exposition, let us assume the number of branches is $1$, for two input vectors $\bx, \by \in \reals^n$, the output vector $\bz \in \reals^n$ is computed as 
\begin{equation}
    \bz = W_3^T \phi\left( (W_1^T x + \bb_1) \otimes (W_2^T y + \bb_2) \right) + \bb_3,
\end{equation}
where $W_1$, $W_2$, $W_3 \in \reals^{n \times n}$ and $\bb_1, \bb_2, \bb_3 \in \reals^n$ are parameters of linear layers, $\phi$ refers to the rectified linear unit (ReLU), and $\otimes$ denotes elementwise product. The implementation for this module in PyTorch is shown in~\cref{list:mbf}.

\section{Modified transformer encoder layer}
\label{app:me}

\begin{table}[t]\small
	\caption{Performance comparison amongst different variants of the cooperative layer on HICO-DET~\cite{hicodet} test set under default setting. All variants below use ResNet50~\cite{resnet} as the backbone CNN and employ one layer. The acronym M.E. stands for modified encoder.}
	\label{tab:coop-variants}
	\setlength{\tabcolsep}{2pt} 
	\vspace{-4pt}
	\begin{tabularx}{\linewidth}{l C C C}
		\toprule
		\textbf{Variant} & \textbf{Full} & \textbf{Rare} & \textbf{Non-rare} \\
        \midrule
        Vanilla & 31.15 $\pm$ .03 & 25.70 &  32.77 \\ 
        Vanilla w/ add. pos. enc. & 31.14 & 25.59 & 32.80 \\
        M.E. w/o pairwise terms & 30.93 & 24.53 & 32.84 \\
        M.E. & \textbf{31.33} $\pm$ .04 & \textbf{26.02} & \textbf{32.91} \\
		\bottomrule
	\end{tabularx}
\end{table}

In this section, we compare the performance and interpretability of the modified transformer encoder layer to alternative formulations. To this end, we test multiple variants of the cooperative layer. As shown in~\cref{tab:coop-variants}, using a vanilla transformer encoder results in a small decrease ($0.2$~mAP) in performance. This gap persists after applying additive positional encodings learned from the unary box terms shown in \cref{eq:unary}. We then demonstrate the importance of the pairwise terms in \cref{eq:pairwise} by removing them from the positional encodings, which resulted in a $0.4$~mAP decrease. Together, these results indicate that the pairwise terms provide useful information for the cooperative layer and a consistent mAP performance boost.

\begin{figure*}[t]
	\begin{subfigure}[t]{0.25\linewidth}
		\centering
		\includegraphics[height=0.69\linewidth]{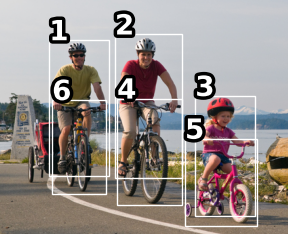}
		\caption{\textit{sample image}}
		\label{fig:sample-image}
	\end{subfigure}
    \hspace{-12pt}
	\begin{subfigure}[t]{0.245\linewidth}
		\centering
		\includegraphics[height=0.8\linewidth]{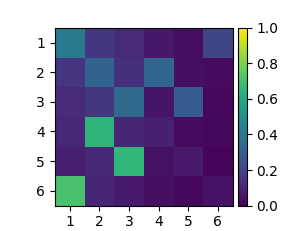}
		\caption{\textit{attn. map from M.E.}}
		\label{fig:attn-me}
	\end{subfigure}
    \hspace{4pt}
	\begin{subfigure}[t]{0.24\linewidth}
		\centering
		\includegraphics[height=0.8\linewidth]{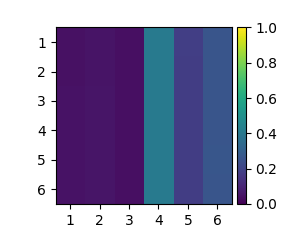}
		\caption{\textit{attn. map from M.E. w/o pairwise terms}}
		\label{fig:attn-me-wo-pairwise}
	\end{subfigure}
    \hspace{5pt}
	\begin{subfigure}[t]{0.24\linewidth}
		\centering
		\includegraphics[height=0.8\linewidth]{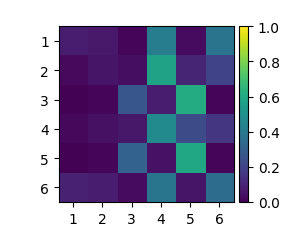}
		\caption{\textit{attn. map from vanilla encoder}}
		\label{fig:attn-vanilla}
	\end{subfigure}
    \caption{An image with detected instances~\subref{fig:sample-image} and attention maps in the cooperative layer with different implementations, including the vanilla encoder~\subref{fig:attn-vanilla}, modified encoder w/o pairwise terms~\subref{fig:attn-me-wo-pairwise} and the modified encoder~\subref{fig:attn-me}.}
    \label{fig:attn-comparison}
\end{figure*}

\begin{figure*}
    \begin{subfigure}[t]{.49\linewidth}
        \centering
        \includegraphics[height=.32\linewidth]{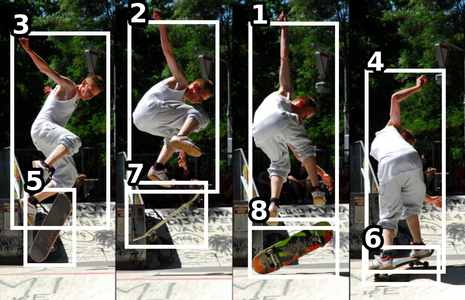}
        \includegraphics[height=.32\linewidth]{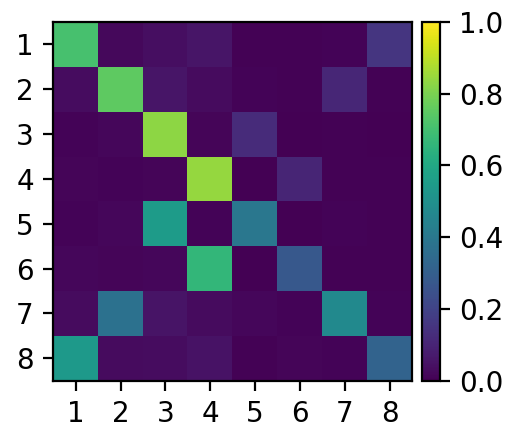}
        \caption{Sample image (left) and the attention map (right) between pairs of unary tokens.}
        \label{fig:sample-unary-attn-1}
    \end{subfigure}
    \quad
    \begin{subfigure}[t]{.49\linewidth}
        \centering
        \includegraphics[height=.32\linewidth]{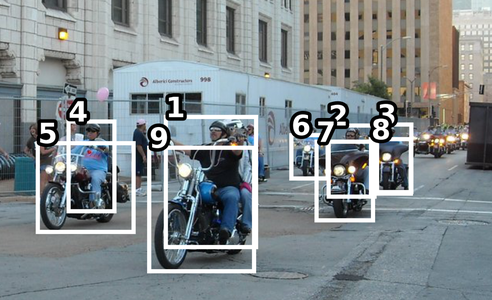}
        \includegraphics[height=.32\linewidth]{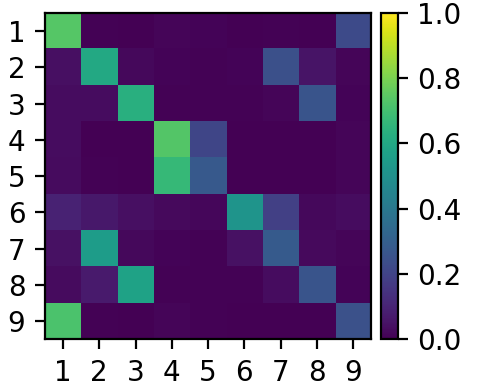}
        \caption{Sample image (left) and the attention map (right) between pairs of unary tokens.}
        \label{fig:sample-unary-attn-2}
    \end{subfigure}

    \begin{subfigure}[t]{.49\linewidth}
        \centering
        \includegraphics[width=.95\linewidth]{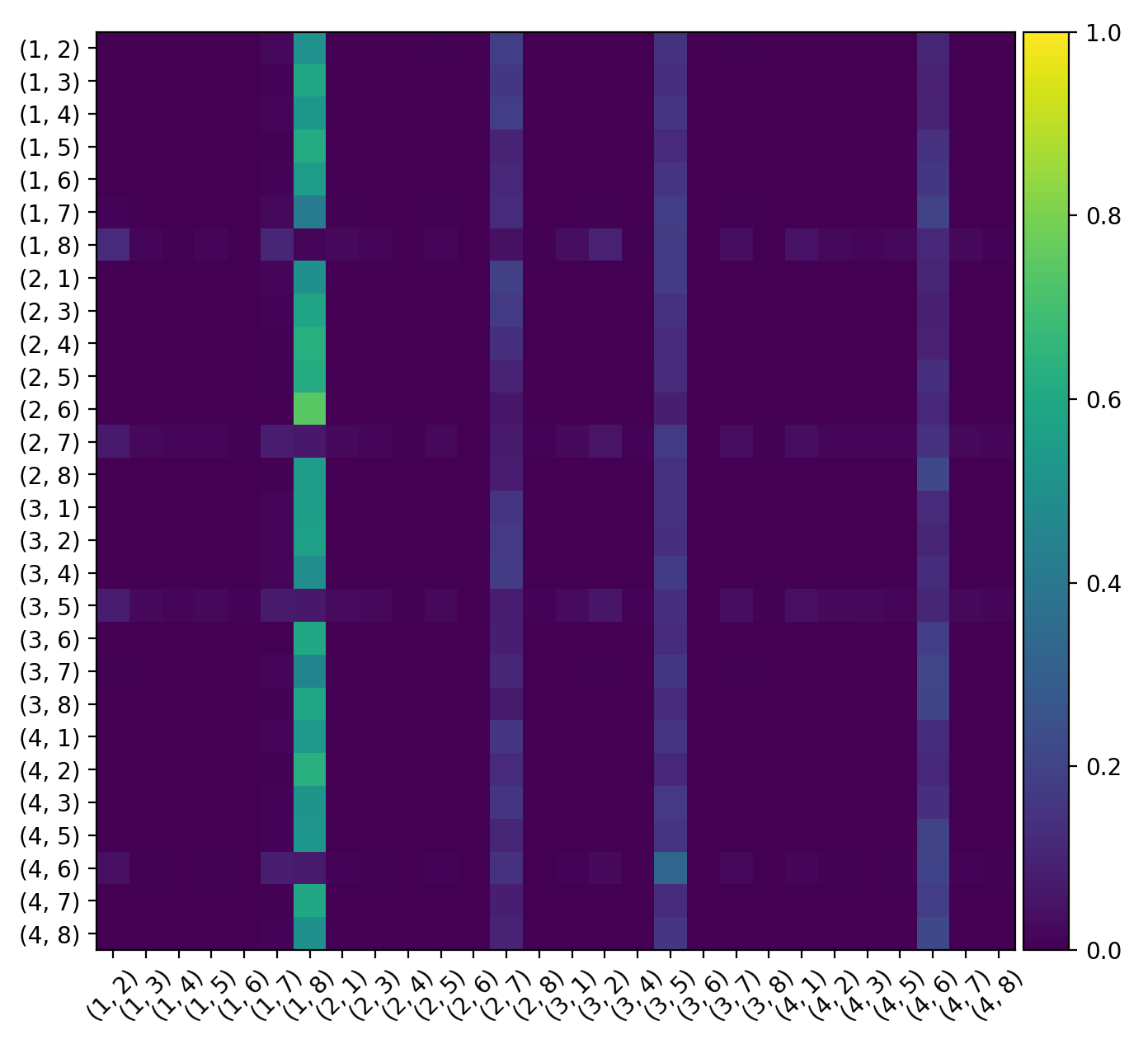}
        \caption{Attention map between pairwise tokens.}
        \label{fig:sample-pairwise-attn-1}
    \end{subfigure}
    \quad
    \begin{subfigure}[t]{.49\linewidth}
        \centering
        \includegraphics[width=.95\linewidth]{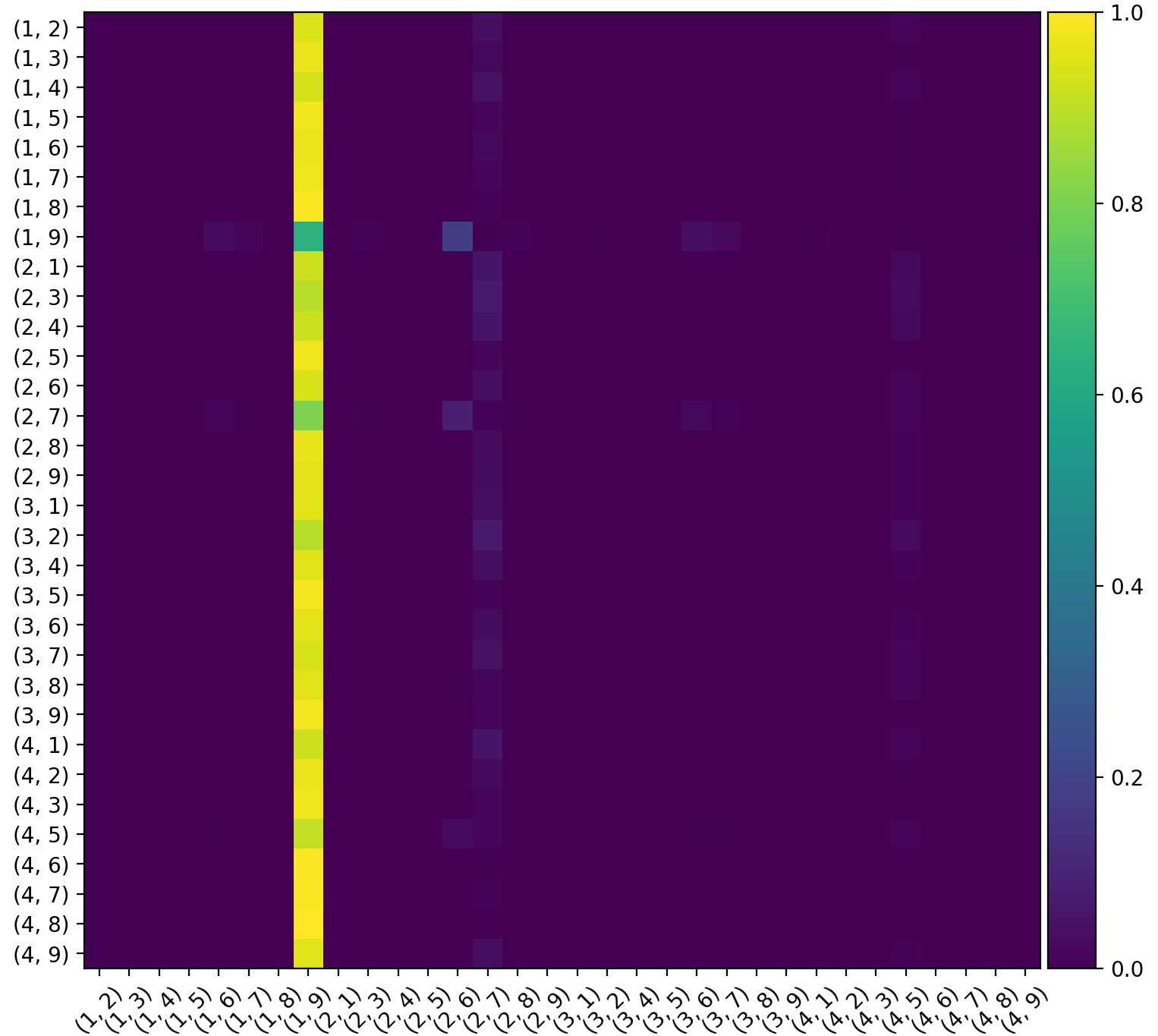}
        \caption{Attention map between pairwise tokens.}
        \label{fig:sample-pairwise-attn-2}
    \end{subfigure}
    \caption{Our model exhibits the same behaviour on sample images with numerous human and object instances. Specifically, the attention map for unary tokens shows high symmetry, where potentially interactive instances attend to each other. And the pairwise attention map indicates that non-interactive pairs attend to the most dominant pairs to be suppressed.}
    \label{fig:add-attn}
    \vspace{-5pt}
\end{figure*}

\begin{lstlisting}[
    float=*t,
    label={list:unary},
    caption={PyTorch implementation of the modified transformer encoder layer. For simplicity, the feedforward network (FFN)~\cite{xfmer} has been omitted, as its architecture and implementation has been publicly available.}
]
import torch
import torch.nn as nn
import torch.nn.functional as F
from typing import Tuple

class ModifiedEncoderLayer(nn.Module):
    def __init__(self, hidden_size: int = 256, repr_size: int = 256, num_heads: int = 8) -> None:
        super().__init__()
        if repr_size % num_heads != 0:
            raise ValueError(
                f"The given representation size {repr_size} "
                f"should be divisible by the number of attention heads {num_heads}.")
        self.sub_repr_size = int(repr_size / num_heads)
        self.hidden_size = hidden_size
        self.repr_size = repr_size
        self.num_heads = num_heads

        self.unary = nn.Linear(hidden_size, repr_size)
        self.pairwise = nn.Linear(repr_size, repr_size)
        self.attn = nn.ModuleList([nn.Linear(3 * self.sub_repr_size, 1) for _ in range(num_heads)])
        self.message = nn.ModuleList([
            nn.Linear(self.sub_repr_size, self.sub_repr_size) for _ in range(num_heads)])
        self.aggregate = nn.Linear(repr_size, hidden_size)
        self.dropout = nn.Dropout(0.1)
        self.norm = nn.LayerNorm(hidden_size)

    def reshape(self, x: torch.Tensor) -> torch.Tensor:
        new_x_shape = x.size()[:-1] + (self.num_heads, self.sub_repr_size)
        x = x.view(*new_x_shape)

        if len(new_x_shape) == 3: return x.permute(1, 0, 2)
        elif len(new_x_shape) == 4: return x.permute(2, 0, 1, 3)
        else: raise ValueError("Incorrect tensor shape")

    def forward(self, x: torch.Tensor, y: torch.Tensor) -> Tuple[torch.Tensor, torch.Tensor]:
        """
        Parameters:
        -----------
        x: torch.Tensor
            Unary tokens of size (N, K)
        y: torch.Tensor
            Pairwise positional encodings of size (N, N, K)
        """
        device = x.device; n = len(x)
        u = F.relu(self.unary(x))
        p = F.relu(self.pairwise(y))
        # Unary features (H, N, K/H)
        u_r = self.reshape(u)
        # Pairwise features (H, N, N, K/H)
        p_r = self.reshape(p)
        # Generate indices for pairwise concatenation
        i, j = torch.meshgrid(torch.arange(n, device=device), torch.arange(n, device=device))
        # Features used to compute attention (H, N, N, 3K/H)
        attn_features = torch.cat([u_r[:, i], u_r[:, j], p_r], dim=-1)
        # Attention weights (H,) (N, N, 1)
        weights = [F.softmax(l(f), dim=0) for f, l in zip(attn_features, self.attn)]
        # Repeated unary feaures along the third dimension (H, N, N, K/H)
        u_r_repeat = u_r.unsqueeze(dim=2).repeat(1, 1, n, 1)
        messages = [l(f_1 * f_2) for f_1, f_2, l in zip(u_r_repeat, p_r, self.message)]
        aggregated_messages = self.aggregate(F.relu(
            torch.cat([(w * m).sum(dim=0) for w, m in zip(weights, messages)], dim=-1)
        ))
        aggregated_messages = self.dropout(aggregated_messages)
        x = self.norm(x + aggregated_messages)
        return x, weights
\end{lstlisting}

In addition, we show the attention maps in different variants of the cooperative layer in~\cref{fig:attn-comparison}. Notably, our modified encoder (\cref{fig:attn-me}) accurately infers the correspondence between instances, where the interactive humans and objects attend to each other. This suggests that the pairwise positional encoding instills an inductive bias in the modified encoder that allows it to identify interactive and non-interaction pairs, and preferentially share information between the interactive ones. Furthermore, we show that, without the pairwise terms, as shown in~\cref{fig:attn-me-wo-pairwise}, the attention map becomes uniform along one dimension. Similarly, the vanilla encoder does not make use of the pairwise spatial information either. This results in the attention maps being much less interpretable as shown in~\cref{fig:attn-vanilla}. In particular, there is no strong mutual attention between interactive instances, but more attention between non-interactive ones. The complete implementation of the modified encoder in PyTorch is shown in~\cref{list:unary}.

\begin{figure*}
    \begin{subfigure}[t]{0.19\linewidth}
        \centering
        \includegraphics[height=0.7\linewidth]{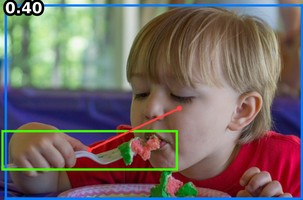}
        \caption{\textit{holding a fork}}
        \label{fig:holding-fork}
    \end{subfigure}
    \hspace{4pt}
    \begin{subfigure}[t]{0.19\linewidth}
        \centering
        \includegraphics[height=0.7\linewidth]{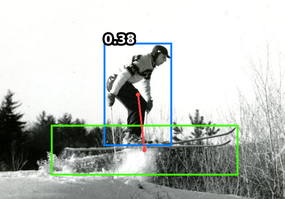}
        \caption{\textit{jumping skis}}
        \label{fig:jumping-ski}
    \end{subfigure}
    \hspace{-1pt}
    \begin{subfigure}[t]{0.19\linewidth}
        \centering
        \includegraphics[height=0.7\linewidth]{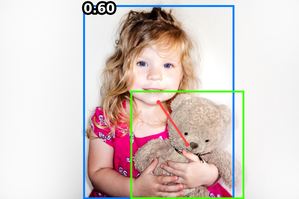}
        \caption{\textit{holding a teddy bear}}
        \label{fig:holding-teddy-bear}
    \end{subfigure}
    \hfill%
    \begin{subfigure}[t]{0.19\linewidth}
        \centering
        \includegraphics[height=0.7\linewidth]{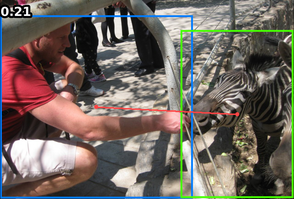}
        \caption{\textit{petting a zebra}}
        \label{fig:petting-zebra}
    \end{subfigure}
    \hfill%
    \begin{subfigure}[t]{0.19\linewidth}
        \centering
        \includegraphics[height=0.7\linewidth]{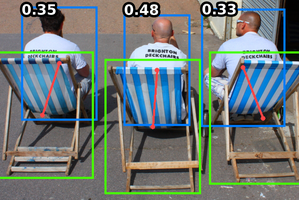}
        \caption{\textit{sitting on a chair}}
        \label{fig:sitting-on-chair}
    \end{subfigure}

    \begin{subfigure}[t]{0.19\linewidth}
        \centering
        \includegraphics[height=0.7\linewidth]{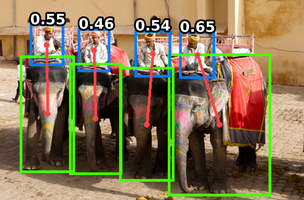}
        \caption{\textit{riding an elephant}}
        \label{fig:riding-elephant}
    \end{subfigure}
    \hspace{4pt}
    \begin{subfigure}[t]{0.19\linewidth}
        \centering
        \includegraphics[height=0.7\linewidth]{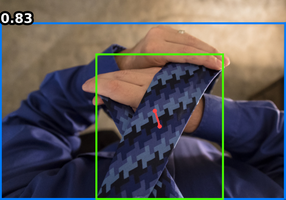}
        \caption{\textit{wearing a tie}}
        \label{fig:wearing-tie}
    \end{subfigure}
    \hspace{-1pt}
    \begin{subfigure}[t]{0.19\linewidth}
        \centering
        \includegraphics[height=0.7\linewidth]{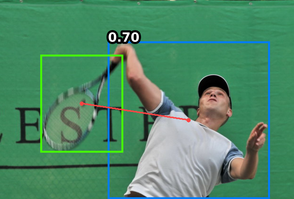}
        \caption{\textit{swinging a tennis racket}}
        \label{fig:swinging-tennis-racket}
    \end{subfigure}
    \hfill%
    \begin{subfigure}[t]{0.19\linewidth}
        \centering
        \includegraphics[height=0.7\linewidth]{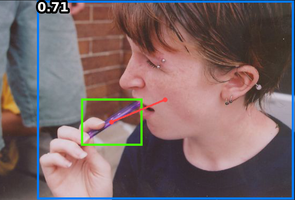}
        \caption{\textit{holding a toothbrush}}
        \label{fig:holding-toothbrush}
    \end{subfigure}
    \hfill%
    \begin{subfigure}[t]{0.19\linewidth}
        \centering
        \includegraphics[height=0.7\linewidth]{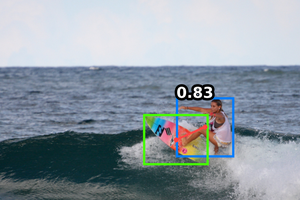}
        \caption{\textit{standing on a surfboard}}
        \label{fig:standing-on-surfboard}
    \end{subfigure}
    \caption{Additional qualitative results for detected human--object pairs on HICO-DET~\cite{hicodet} test set.}
    \label{fig:qual-hicodet}
\end{figure*}

\begin{figure*}
    \begin{subfigure}[t]{0.19\linewidth}
        \centering
        \includegraphics[height=0.7\linewidth]{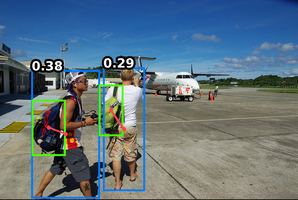}
        \caption{\textit{carrying a backpack}}
        \label{fig:vcoco-carrying-backpack}
    \end{subfigure}
    \hfill%
    \begin{subfigure}[t]{0.19\linewidth}
        \centering
        \includegraphics[height=0.7\linewidth]{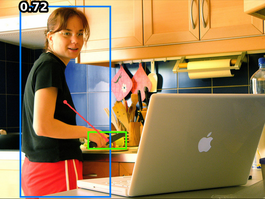}
        \caption{\textit{cutting with a knife}}
        \label{fig:vcoco-cutting-knife}
    \end{subfigure}
    \begin{subfigure}[t]{0.19\linewidth}
        \centering
        \includegraphics[height=0.7\linewidth]{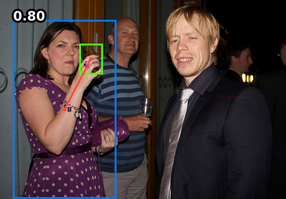}
        \caption{\textit{drinking from a bottle}}
        \label{fig:vcoco-drinking-bottle}
    \end{subfigure}
    \hfill%
    \begin{subfigure}[t]{0.19\linewidth}
        \centering
        \includegraphics[height=0.7\linewidth]{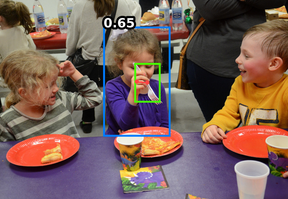}
        \caption{\textit{eating with a fork}}
        \label{fig:vcoco-eating-fork}
    \end{subfigure}
    \hfill%
    \begin{subfigure}[t]{0.19\linewidth}
        \centering
        \includegraphics[height=0.7\linewidth]{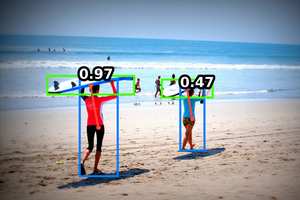}
        \caption{\textit{holding a surfboard}}
        \label{fig:vcoco-holding-surfboard}
    \end{subfigure}

    \begin{subfigure}[t]{0.19\linewidth}
        \centering
        \includegraphics[height=0.7\linewidth]{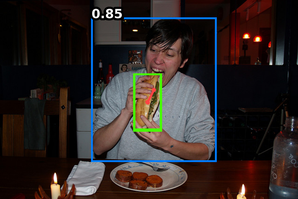}
        \caption{\textit{eating a hotdog}}
        \label{fig:vcoco-eating-hotdog}
    \end{subfigure}
    \hfill%
    \begin{subfigure}[t]{0.19\linewidth}
        \centering
        \includegraphics[height=0.7\linewidth]{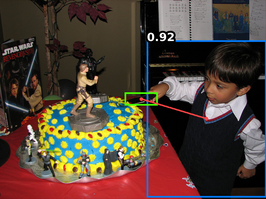}
        \caption{\textit{holding a knife}}
        \label{fig:vcoco-holding-knife}
    \end{subfigure}
    \begin{subfigure}[t]{0.19\linewidth}
        \centering
        \includegraphics[height=0.7\linewidth]{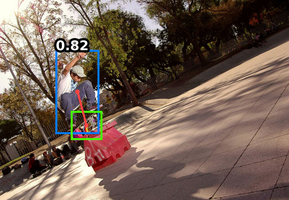}
        \caption{\textit{jumping a skateboard}}
        \label{fig:vcoco-jumping-stakeboard}
    \end{subfigure}
    \hfill%
    \begin{subfigure}[t]{0.19\linewidth}
        \centering
        \includegraphics[height=0.7\linewidth]{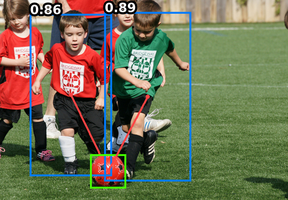}
        \caption{\textit{kicking a sports ball}}
        \label{fig:vcoco-kicking-sportsball}
    \end{subfigure}
    \hfill%
    \begin{subfigure}[t]{0.19\linewidth}
        \centering
        \includegraphics[height=0.7\linewidth]{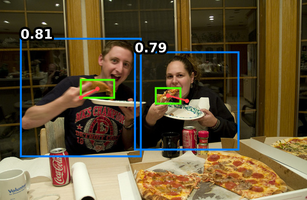}
        \caption{\textit{eating a pizza}}
        \label{fig:vcoco-eating-pizza}
    \end{subfigure}

    \begin{subfigure}[t]{0.19\linewidth}
        \centering
        \includegraphics[height=0.7\linewidth]{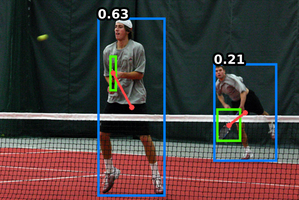}
        \caption{\textit{holding a tennis racket}}
        \label{fig:vcoco-holding-tennis-racket}
    \end{subfigure}
    \hfill%
    \begin{subfigure}[t]{0.19\linewidth}
        \centering
        \includegraphics[height=0.7\linewidth]{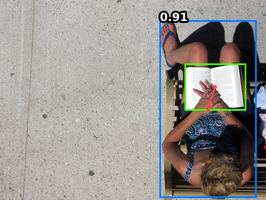}
        \caption{\textit{reading a book}}
        \label{fig:vcoco-reading-book}
    \end{subfigure}
    \begin{subfigure}[t]{0.19\linewidth}
        \centering
        \includegraphics[height=0.7\linewidth]{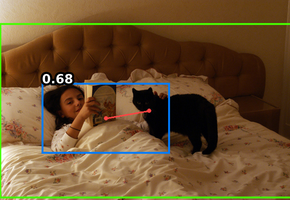}
        \caption{\textit{laying on a bed}}
        \label{fig:vcoco-laying-bed}
    \end{subfigure}
    \hfill%
    \begin{subfigure}[t]{0.19\linewidth}
        \centering
        \includegraphics[height=0.7\linewidth]{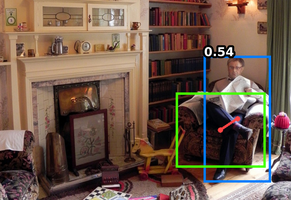}
        \caption{\textit{sitting on a couch}}
        \label{fig:vcoco-sitting-couch}
    \end{subfigure}
    \hfill%
    \begin{subfigure}[t]{0.19\linewidth}
        \centering
        \includegraphics[height=0.7\linewidth]{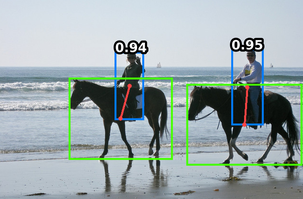}
        \caption{\textit{riding a horse}}
        \label{fig:vcoco-riding-horse}
    \end{subfigure}

    \begin{subfigure}[t]{0.19\linewidth}
        \centering
        \includegraphics[height=0.7\linewidth]{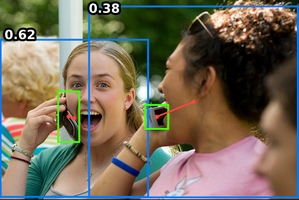}
        \caption{\textit{talking on a phone}}
        \label{fig:vcoco-talking-phone}
    \end{subfigure}
    \hfill%
    \begin{subfigure}[t]{0.19\linewidth}
        \centering
        \includegraphics[height=0.7\linewidth]{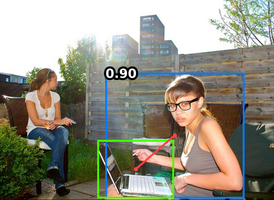}
        \caption{\textit{working on a computer}}
        \label{fig:vcoco-working-computer}
    \end{subfigure}
    \begin{subfigure}[t]{0.19\linewidth}
        \centering
        \includegraphics[height=0.7\linewidth]{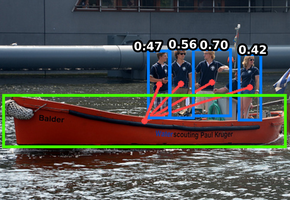}
        \caption{\textit{riding a boat}}
        \label{fig:vcoco-riding-boat}
    \end{subfigure}
    \hfill%
    \begin{subfigure}[t]{0.19\linewidth}
        \centering
        \includegraphics[height=0.7\linewidth]{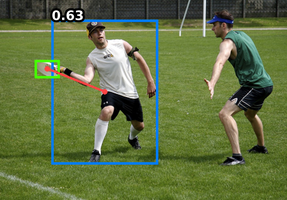}
        \caption{\textit{throwing a frisbee}}
        \label{fig:vcoco-throwing-frisbee}
    \end{subfigure}
    \hfill%
    \begin{subfigure}[t]{0.19\linewidth}
        \centering
        \includegraphics[height=0.7\linewidth]{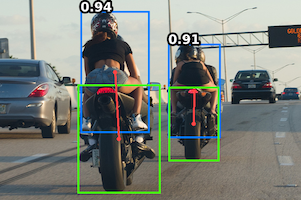}
        \caption{\textit{riding a motorcycle}}
        \label{fig:vcoco-riding-motorcycle}
    \end{subfigure}

    \caption{Additional qualitative results for detected human--object pairs on V-COCO~\cite{vcoco} test set.}
    \label{fig:qual-vcoco}
\end{figure*}

\begin{figure*}
    \centering
    \begin{subfigure}[t]{0.19\linewidth}
        \includegraphics[height=0.7\linewidth]{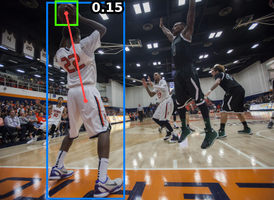}
        \caption{\textit{blocking a sports ball}}
        \label{fig:fp-blocking-ball}
    \end{subfigure}
    \hspace{-3pt}
    \begin{subfigure}[t]{0.19\linewidth}
        \includegraphics[height=0.7\linewidth]{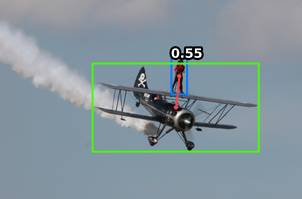}
        \caption{\textit{flying an airplane}}
        \label{fig:fp-flying-airplane}
    \end{subfigure}
    \hspace{5pt}
    \begin{subfigure}[t]{0.19\linewidth}
        \includegraphics[height=0.7\linewidth]{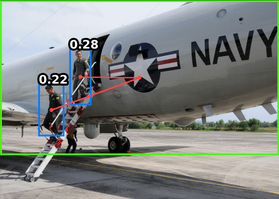}
        \caption{\textit{boarding an airplane}}
        \label{fig:fp-boarding-airplane}
    \end{subfigure}
    \begin{subfigure}[t]{0.19\linewidth}
        \includegraphics[height=0.7\linewidth]{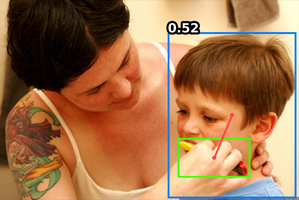}
        \caption{\textit{brushing with a toothbrush}}
        \label{fig:fp-brushing-toothbrushl}
    \end{subfigure}
    \hspace{5pt}
    \begin{subfigure}[t]{0.19\linewidth}
        \includegraphics[height=0.7\linewidth]{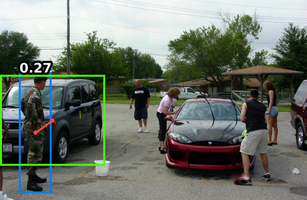}
        \caption{\textit{washing a car}}
        \label{fig:fp-washing-car}
    \end{subfigure}
    \caption{False positives on HICO-DET~\cite{hicodet} test set.}
    \label{fig:fp}
    \vspace{-5pt}
\end{figure*}

\section{Additional qualitative results}
\label{app:qual}

We show more visualisations of attention maps in~\cref{fig:add-attn}. We intentionally avoided images with very few human and object instances and instead selected those with more complicated scenes for the purpose of demonstration. As shown by the attention maps, our model behaves consistently across different interaction types. More qualitative results for detected HOIs from HICO-DET~\cite{hicodet} and V-COCO~\cite{vcoco} can be found in~\cref{fig:qual-hicodet} and~\cref{fig:qual-vcoco} respectively.

To better understand the limitations of our model, we also show some false positives on HICO-DET in~\cref{fig:fp}. In particular, the model sometimes struggles to identify the correct human instance for the interaction as shown in~\cref{sub@fig:fp-blocking-ball,sub@fig:fp-flying-airplane,sub@fig:fp-brushing-toothbrushl}. This is largely due to the plausible spatial relationship and the saliency of the human instance, both of which the model relies on heavily. Another false positive of similar cause can be seen in~\cref{fig:fp-boarding-airplane}, where both human instances are plausible candidates for the interaction \textit{boarding airplane} based on their spatial locations.

\section{Asset attrition}
Annotations from the HICO-DET~\cite{hicodet} dataset have no license specified. Images from this dataset are licensed under Creative Commons from Flickr. Annotations from the V-COCO~\cite{vcoco} dataset are licensed under the MIT License. V-COCO makes use of annotations and images from the MS COCO~\cite{coco} dataset. The MS-COCO annotations are licensed under a Creative Commons Attribution 4.0 License, and the images are sourced from from Flickr and have a variety of Creative Commons licenses, listed in the MS-COCO annotation files: Attribution-NonCommercial-ShareAlike License, Attribution-NonCommercial License, Attribution-NonCommercial-NoDerivs License, Attribution License, Attribution-ShareAlike License, Attribution-NoDerivs License, and No known copyright restrictions.

\end{document}